\documentclass[10pt,twocolumn,letterpaper]{article}

\usepackage{cvpr}      
\usepackage{kotex}  
\usepackage{newtxtext}
\usepackage[table,dvipsnames,svgnames,x11names]{xcolor}
\usepackage{colortbl}       
\usepackage{tcolorbox}
\usepackage{enumitem}
\usepackage{booktabs}
\usepackage{subcaption}
\usepackage{tabularx}
\usepackage{siunitx}
\usepackage{graphicx}
\usepackage{float}
\usepackage[export]{adjustbox}
\usepackage{multirow}
\usepackage{listings}
\usepackage[T1]{fontenc}
\usepackage[pagebackref,breaklinks,colorlinks,allcolors=cvprblue]{hyperref}
\definecolor{promptbg}{rgb}{0.95,0.95,0.95}
\definecolor{cvprblue}{rgb}{0.21,0.49,0.74}
\usepackage{titletoc}

\newcolumntype{Y}{>{\raggedright\arraybackslash}X}

\newcommand{\task}{\textbf{\textsc{CultureMix}}}

\newcommand{\SF}{\texttt{SF}\xspace}
\newcommand{\MF}{\texttt{MF}\xspace}
\newcommand{\SFB}{\texttt{SFB}\xspace}
\newcommand{\MFB}{\texttt{MFB}\xspace}

\newcommand{\lightmidrule}{\arrayrulecolor{black!40}\midrule\arrayrulecolor{black}}

\def\paperID{3550} 
\def\confName{CVPR}
\def\confYear{2026}

\title{\textsc{World in a Frame}:\\ Understanding Culture Mixing as a New Challenge for Vision-Language Models}

\author{%
Eunsu Kim$^{1,*}$ ,
Junyeong Park$^{1,*}$ ,
  Na Min An$^{1,*}$, 
  Junseong Kim$^{1,\dagger}$, 
 Hitesh Laxmichand Patel$^{2,\dagger}$, \\Jiho Jin$^{1,\dagger}$, Julia Kruk$^3$, Amit Agarwal$^2$, Srikant Panda$^2$, Fenal Ashokbhai Ilasariya$^4$, \\Hyunjung Shim$^{1,\ddagger}$, 
  {Alice Oh}$^{1,\ddagger}$ \\
  $^1$KAIST, $^2$Oracle, $^3$Meta, $^4$Stevens Institute of Technology
\\
\texttt{\{kes0317, jjjunyeong9986, naminan\}@kaist.ac.kr} \\ 
}

\begin{document}
\maketitle
\begin{abstract}

In a globalized world, cultural elements from diverse origins frequently appear together within a single visual scene. We refer to these as culture mixing scenarios, yet how Large Vision-Language Models (LVLMs) perceive them remains underexplored. 
We investigate culture mixing as a critical challenge for LVLMs and examine how current models behave when cultural items from multiple regions appear together. To systematically analyze these behaviors, we construct \task, a food Visual Question Answering (VQA) benchmark with 23k diffusion-generated, human-verified culture mixing images across four subtasks: (1) food-only, (2) food+food, (3) food+background, and (4) food+food+background.
Evaluating 10 LVLMs, we find consistent failures to preserve individual cultural identities in mixed settings. Models show strong background reliance, with accuracy dropping 14\% when cultural backgrounds are added to food-only baselines, and they produce inconsistent predictions for identical foods across different contexts.
To address these limitations, we explore three robustness strategies. We find supervised fine-tuning using a diverse culture mixing dataset substantially improve model consistency and reduce background sensitivity.
We call for increased attention to culture mixing scenarios as a critical step toward developing LVLMs capable of operating reliably in culturally diverse real-world environments.


\begin{adjustbox}{width=\linewidth}

\begin{tabular}{@{}c@{\hspace{6pt}}c@{}}
\centering
  \raisebox{-0.25\height}{\includegraphics[height=12pt]{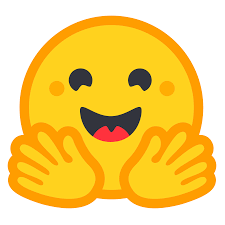}} &
  \href{https://huggingface.co/datasets/EunsuKim/CultureMix}{\centering{\texttt{huggingface.co/datasets/EunsuKim/CultureMix}}} \\
\end{tabular}
\end{adjustbox}

\end{abstract}    
\section{Introduction}
\label{sec:intro}
\vspace{-4mm}
\noindent\begin{figure}[t]
    \centering
    \includegraphics[width=0.78\columnwidth]{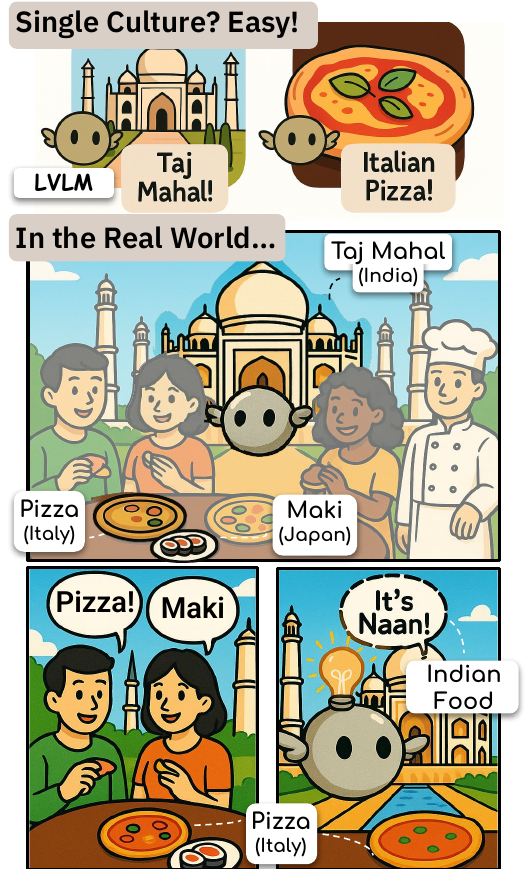}\caption{\textbf{Conceptual illustration of LVLMs in Culture Mixing scenarios.} Real-world contexts often contain multiple cultural elements that humans can easily identify, yet LVLMs struggle to identify them.\protect\footnotemark}
    \label{fig:teaser}
    \vspace{-1em}
\end{figure}
\begingroup
\renewcommand\thefootnote{}%
\footnote{$*$,$\dagger$ denote equal contributions, $\ddagger$ Senior Authors.}%
\addtocounter{footnote}{-1}%
\endgroup
\footnotetext{The figure was generated with assistance from ChatGPT.}

\noindent\begin{figure*}[t]
    \centering
    \includegraphics[width=\textwidth]{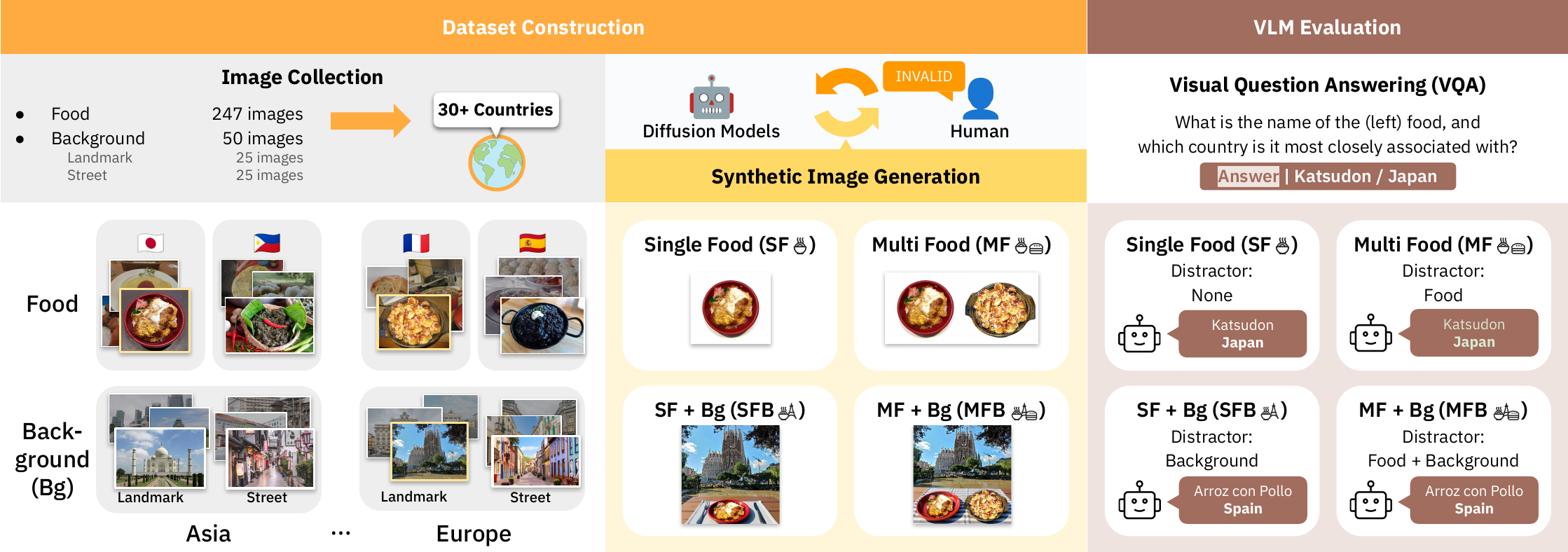}
    \caption{\textbf{Dataset construction and evaluation pipeline.} \task aims to benchmark state-of-the-art LVLMs on their cultural knowledge in diverse mixing scenarios by asking models to identify food names and their countries of origin, featuring various combinations of foods and backgrounds from over 30 countries. All images are synthetically generated using diffusion models with human-in-the-loop validation. Model responses in this figure are from InternVL2-14B.}
    \label{fig:pipeline}
\end{figure*}

In today's globalized society, cultures are blending and being shared more than ever. For instance, we often see food serving as a medium where cultures meet, from diverse cuisines in buffets to ramen shops near the Eiffel Tower. This phenomenon, referred to as \textbf{culture mixing}~\cite{hao2016advancing}, captures the coexistence and fusion of multiple cultural cues within a single scene, spanning objects, foods, and environments (see Figure~\ref{fig:teaser}). Nevertheless, each element retains its distinct cultural identity regardless of its location or the cultural elements that coexist with~\cite{Martin2019-qd,Ye2021CultureMixing,Chua2023CultureMixingLeadership}. 
The ability to recognize and preserve these individual cultural identities within mixed contexts is essential for maintaining authenticity and preventing the dilution or misrepresentation of any culture.

Meanwhile, as LVLMs increasingly interact with humans across diverse cultural contexts, recent studies have explored cultural understanding through Visual Question Answering (VQA) benchmarks~\citep{burda2025culturally, liu2025culturevlm, yadav2025beyond}. However, these benchmarks depict scenes rooted in a single cultural context, ignoring culture mixing scenarios. Since culture mixing poses a unique challenge requiring models to distinguish and integrate multiple conflicting cultural cues, models trained on single-culture data may not generalize effectively to such contexts.

To this end, we define culture mixing as a new challenge for LVLMs and evaluate how current models behave under such scenarios. We introduce \task, a systematic benchmark dataset designed to reveal how LVLMs recognize cultural elements at different levels of culturally mixed contexts (Figure \ref{fig:pipeline}). Our large-scale dataset consists of 23k synthetic images (featuring 247 unique food items) spanning 30 countries and 50 unique background seed images, along with 100 real-world images (featuring 219 unique food items). It adopts a VQA format in which models identify a target food item and determine its cultural origin in the presence of other cultural elements--referred to as \textit{cultural distractors}. Depending on the types of these distractors--none, food, background, or both--our dataset enables a structured analysis of how different cultural cues and their combinations influence model behavior. Additionally, it consists of multiple levels of cultural distance between the target item and its distractors, allowing a quantitative examination of the effect of cultural distance on model performance.

 
With \task, we evaluate 10 LVLMs and uncover critical limitations in culturally mixed scenarios, along with key insights to improve model robustness. Across all subtasks, performance decreases noticeably compared to single-culture images, with declines of 1-8\%p in food identification and 1-14\%p in cultural origin prediction. These declines further intensify as the cultural distance between the target and distractor elements increases. 
Our analysis reveals that background cues exert a stronger influence than food distractors, frequently shifting predictions toward the distractor's cultural source and undermining output consistency. This pattern indicates that LVLMs rely heavily on contextual signals rather than the target object itself, and that their predictions are heavily influenced by the cultural relationship between the target and its surrounding context. This issue becomes particularly problematic when visual cues conflict (\ie, co-existing objects from culturally distant regions), often leading to misidentification and cultural bias toward the culture with which the model is more familiar.

Motivated by these findings, we explore both training-free and training-based (SFT) methods to improve model robustness in culturally mixed settings. Both approaches yield consistent gains in accuracy and consistency, highlighting promising directions to enhance cross-cultural understanding while leaving open the question of optimal training objectives for culture mixing scenarios.

In summary, our contributions are as follows:
\begin{itemize}
    \item \textbf{Task.} We introduce \task, a benchmark for systematically evaluating LVLMs' cross-cultural understanding in culturally mixed contexts.
    \item \textbf{Dataset and Resources.} We build and release 23k synthetic images using diffusion models that blends cultural elements from diverse regions.
    \item \textbf{Analysis.} Evaluating 10 LVLMs, we find that existing models struggle to interpret culturally mixed scenes, with accuracy decreasing as cultural distance increases.
    \item \textbf{Mitigation.} We explore training-free and training-based methods to improve model robustness, showing consistent gains while highlighting remaining challenges in designing optimal training objectives for culture mixing.
\end{itemize}

\section{\task}
\label{sec:pipeline}

We propose \task, a novel and challenging task to evaluate cross-cultural awareness of LVLMs through the concept of culture mixing.

In this section, we outline the task (\S~\ref{sec:dataset_task}), the dataset construction pipeline (\S~\ref{sec:dataset_construction}), and dataset statistics (\S~\ref{sec:dataset_stats}).

\subsection{Task Overview}
\label{sec:dataset_task}

We formulate our task as a VQA problem in which models infer both the \textit{food name} and the \textit{country of origin} of a target food item. Each target food item is evaluated under four distinct subtasks that systematically vary the type of cultural distractors—visual cultural elements that co-occur with the target food item in the image. The conditions include: no distractors, only food-type distractors, only background-type distractors, or both types combined.
The four subtasks are defined as follows:
\begin{itemize}[leftmargin=2em]
    \item \textbf{Single Food (\SF} \raisebox{-0.1em}{\includegraphics[height=1em]{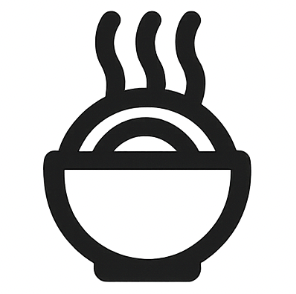}} \textbf{)} \\
    An image of a single dish without any cultural distractors.
This subtask isolates the target food item.

    \item \textbf{Multiple Foods (\MF \raisebox{-0.1em}{\includegraphics[height=1em]{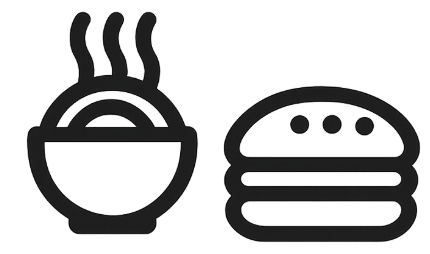} \textbf{)}}}\\
    An image containing multiple dishes, introducing \textit{food-type distractors}.  

    \item \textbf{Single Food with Background (\SFB \raisebox{-0.1em}{\includegraphics[height=1.2em]{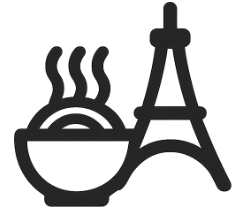}} \textbf{)}} \\
    An image of a single dish accompanied by \textit{background-type cultural distractors}, that reflect a specific cultural context. 
    \item \textbf{Multiple Foods with Background (\MFB} \raisebox{-0.2em}{\includegraphics[height=1.3em]{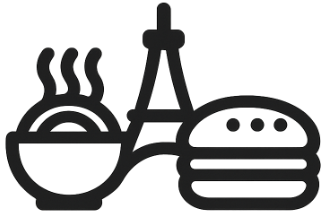}} \textbf{)} \\
    An image containing multiple dishes together with culturally informative backgrounds, containing \textit{both food and background distractors}.  
\end{itemize}

Using \SF as a baseline, our task design reveals how the model's behavior shifts when distractors from different types and diverse countries are introduced. Specifically, we ensure diverse pairings between target foods and distractors in terms of (1) country combinations and (2) cultural distance (See \S~\ref{sec:dataset_stats} for details).

\subsection{Dataset Construction Pipeline}
\label{sec:dataset_construction}

We aim to construct a dataset where distractors are systematically introduced, aligned with our four-subtask design.
We construct a synthetic dataset using editing-based text-to-image diffusion models (FLUX.1-Kontext~\citep{labs2025flux1kontextflowmatching} and Qwen-Image-Edit~\citep{wu2025qwen}). We use existing datasets annotated with the names of foods and their countries of origin as a seed dataset. A rigorous human-in-the-loop pipeline supports the entire process.

\vspace{0.5em}
\noindent\textbf{Seed Dataset Collection.}
We sample food and background images from existing datasets: For foods, we collected 247 seed food images from existing multicultural VQA datasets: WorldCuisines~\cite{winata-etal-2025-worldcuisines} and WorldWideDishes~\cite{10.1145/3715275.3732019}. Based on the availability in existing datasets, we selected 30 countries that each 
have at least 10 food images in our seed sources. From this, we select countries to ensure balanced representation 
across four continents and across their cultural resource levels. Since
resource levels for images are not available, we use language resource levels as 
a proxy, including both high-resource (\eg, United States, United Kingdom, and China) and low-resource  (\eg, Philippines, Algeria, and Croatia).
We also sample 50 seed background images from existing datasets: 
landmark images from VIPPGeo~\cite{10094908} and street images from the Google Landmarks Dataset v2~\cite{9157053}.
We include five images from each of the five continents--the same four continents as in the food list, with the addition of South America.
The complete list of countries and their statistics is provided in the Appendix~\ref{appendix:stats}.\footnote{When suitable images were unavailable, we manually collected additional ones via Google Image Search with clear textual labels in the accompanying metadata.}

\begin{table}[t]
    \centering
    \caption{%
        \textbf{Dataset composition and statistics.}
        Background (\texttt{BG}) includes street and landmark images.
        In \MFB, a single background image from each continent is randomly
        selected and combined due to the large number of possible combinations.
    }
    \resizebox{\linewidth}{!}{%
        \centering
\resizebox{\linewidth}{!}{%
\begin{tabular}{@{} >{\ttfamily}lllr}
\toprule
\textbf{Type} & \textbf{Description} & \textbf{Composition} & \textbf{Size} \\
\midrule
Food & Food & 30 countries, 4 continents & 247 \\ 
BG & Background & n countries, 5 continents, 2 types (landmark, street) & 50 \\ \lightmidrule
\SF & Food & \texttt{Food} +  \texttt{Food} $\times$ 3 Data Augmentation & 988 \\ 
\MF & Food + Food & Food pairs from \SF & 948 \\
\SFB & Food + BG & \SF $\times$ (5 continents $\times$ $\{5$ landmark$,~5$ street$\}$) & 12,350 \\
\MFB & Food + Food + BG & \MF $\times$ (5 continents $\times$ $\{1$ landmark$,~1$ street$\}$) & 9,480 \\
\bottomrule
\end{tabular}%
}

    }
    \label{tab:dataset-stats}
\end{table}

\begin{figure*}[ht]
    \centering
    \begin{subfigure}[t]{0.48\textwidth}
        \centering
        \includegraphics[width=\textwidth]{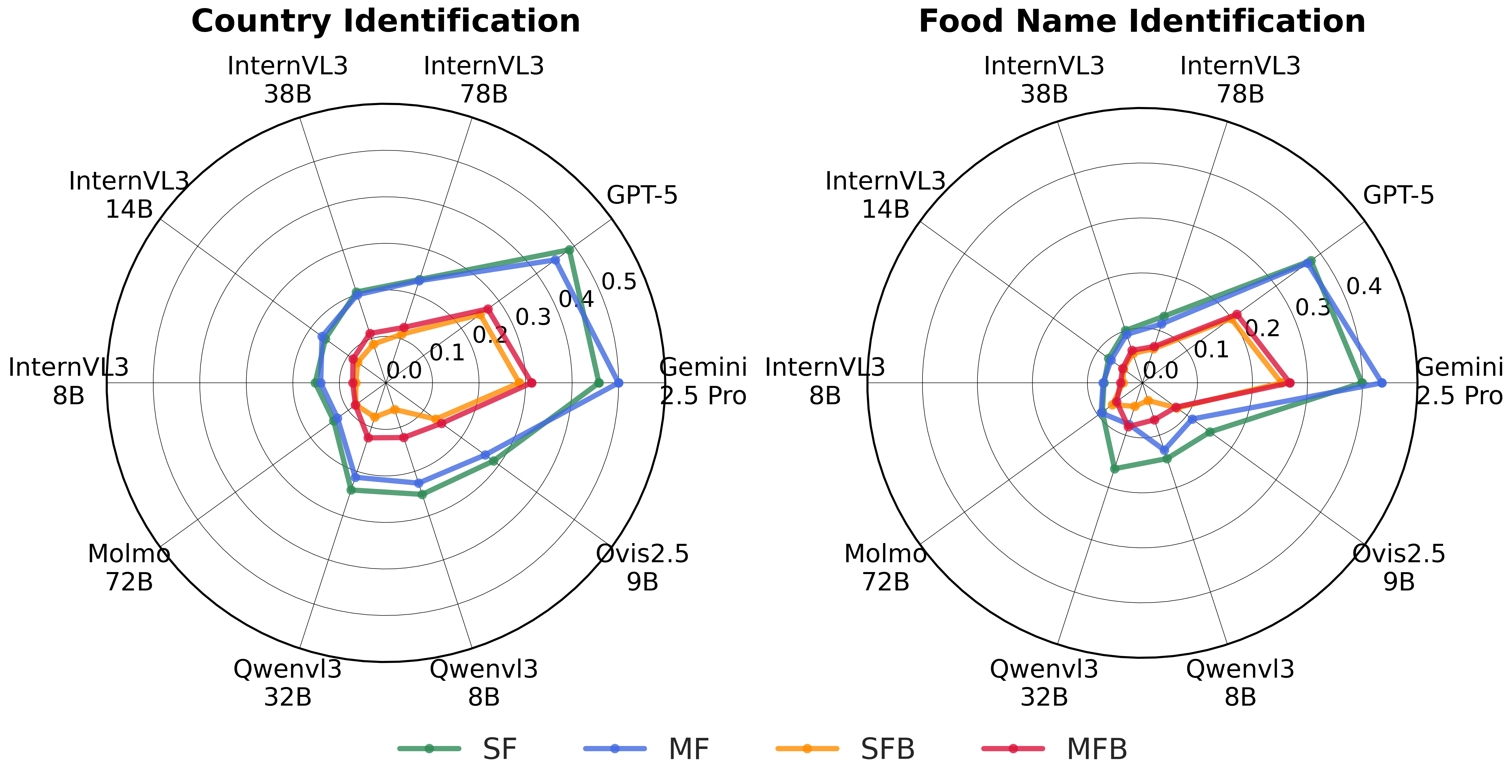}
        \caption{Country and food name identification accuracy.}
        \label{fig:main_models_acc}
    \end{subfigure}
    \hfill
    \begin{subfigure}[t]{0.48\textwidth}
        \centering
        \includegraphics[width=\textwidth]{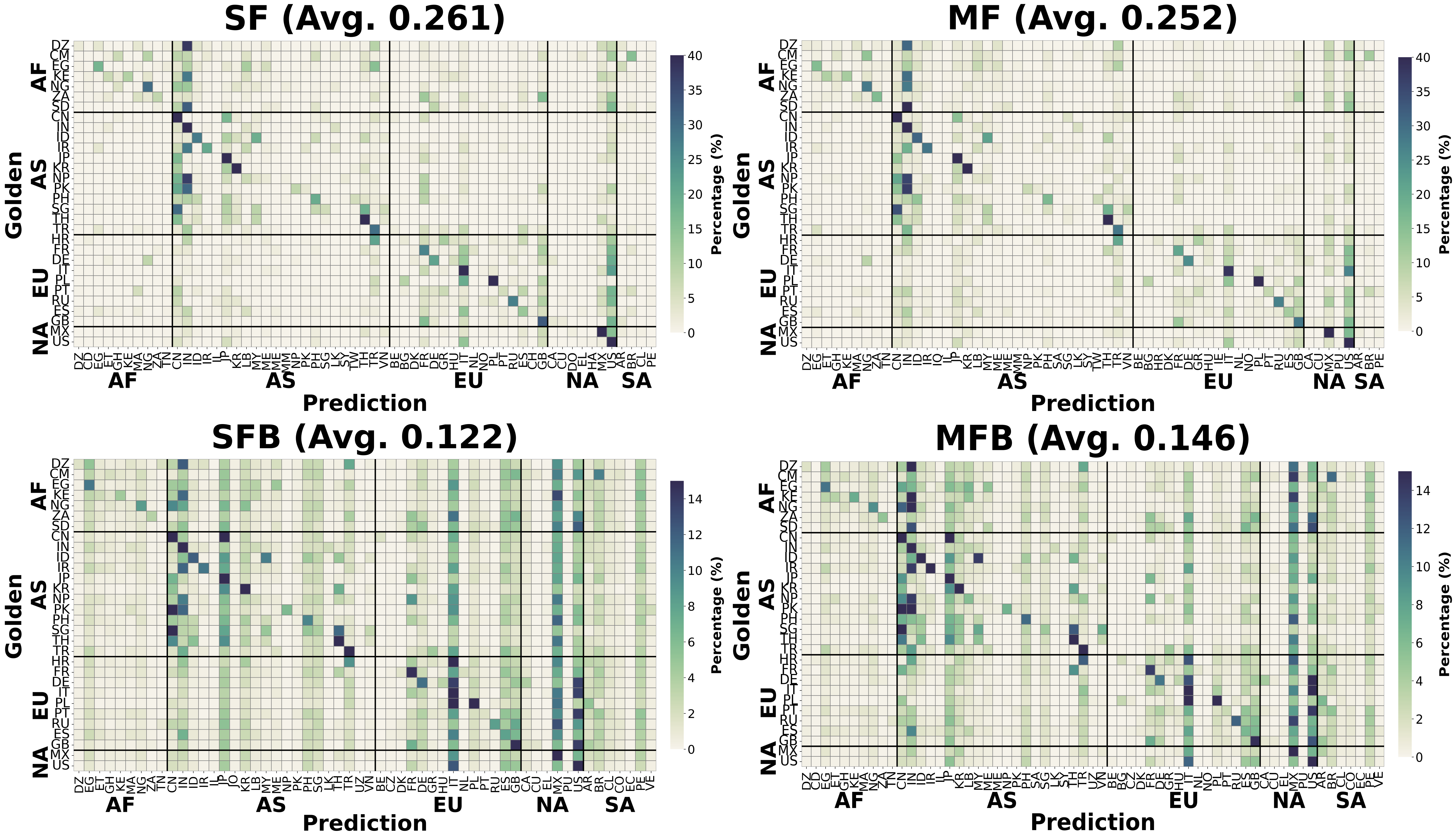}
        \caption{Country identification predictions.}
        \label{fig:main_models_heatmap}
    \end{subfigure}
    \vspace{-4pt}
    \caption{\textbf{Overall model performance on country and food name identification.} (a) Accuracy comparison across models for each subtask. (b) Country identification target-prediction heatmaps for each subtask. For every golden country, the plots show the distribution of predicted countries, illustrating both correct predictions and systematic confusions across models. }
    \vspace{-4pt}
    \label{fig:main_models_acc_heatmap}
\end{figure*}

\vspace{0.5em}
\noindent\textbf{\SF \raisebox{-0.1em}{\includegraphics[height=1.0em]{images/sf.png}}, \MF \raisebox{-0.15em}{\includegraphics[height=1.1em]{images/mf.png}}.}
Using the images from the seed dataset, we remove the original background and replace it with a plain white background to eliminate any additional elements (\eg, text, humans, tables, or scenery) that could unintentionally influence the model’s inference. For this process, we generate candidate images using both FLUX.1-Kontext and Qwen-Image-Edit. Among the two outputs, one of the authors manually selects the image that best preserves fidelity to the original.
The resulting images constitute our \SF set.
We then concatenate two \SF images to compose each \MF image.

\vspace{0.5em}
\noindent\textbf{\SFB \raisebox{-0.1em}{\includegraphics[height=1em]{images/sfb.png}}, \MFB \raisebox{-0.1em}{\includegraphics[height=1.02em]{images/mfb.png}}.}
\SFB and \MFB are \textit{close-up food images with visible backgrounds}.
We first vertically concatenate a background image with \SF and \MF to form \SFB and \MFB, respectively. 
We then apply a diffusion model for image harmonization. We initially start with FLUX.1-Kontext, and when it does not yield satisfactory results, we apply both FLUX.1-Kontext and Qwen-Image-Edit multiple times until the generated images meet our criteria.\footnote{We provide the input examples used for the diffusion models and examples of errored images in Appendix~\ref{appendix:generation_process}, and background images, images of \SF, \MFB, and \task-real samples in Appendix~\ref{appendix:sample_imgs}.}

\vspace{0.5em}

\noindent\textbf{Verification.}
All generation steps were conducted under multiple rounds of human verification until the entire dataset was validated. For each subtask, we established a set of criteria, and any images that did not satisfy the criteria were either regenerated or removed to ensure the quality. We provide a description of the human verification procedure, including image selection criteria, and statistics of the filtered images in Appendix~\ref{appendix:pipeline-details}.

\subsection{Dataset Statistics}
\label{sec:dataset_stats}

We illustrate the procedure for selecting the seed images and constructing their 
combinations for each subset. 
Table~\ref{tab:dataset-stats} summarizes the main statistics for these subsets.

\vspace{-1em}
\paragraph{Food Combinations.}

For every food item, we aim to construct combinations that reflect varying levels of cultural distance. 
We operationalize cultural distance through geographic proximity, creating three levels as follows \cite{Li2023journalsInternational}: 
(i) target and distractor from the same country, 
(ii) from different countries within the same continent, 
and (iii) from different continents. 
In addition, we account for single-image classification performance 
across three baseline models (Qwen2.5-VL-72B-Instruct, GPT-image, and Gemini-2.5-flash). We include both easy cases (all three models correct) and hard 
cases (all three models incorrect), from cases where all three models are correct to cases where all fail, enabling 
analysis of culture mixing effects at varying difficulty levels.
Following these, we construct 948 image pairs spanning 30 countries. 
\vspace{-0.5em}
\paragraph{Background \(\times\) Food Combinations.}  
For \SFB, each of the 247 food items is paired with 10 background combinations (2 types × 5 continents), with 5 images per continent, yielding a total of \textbf{12,350} images.
For \MFB, each of the 948 food image pairs is combined with randomly sampled backgrounds from each continent, resulting in \textbf{9,480} images.

\section{Experiments}
\label{sec:experiments}

\subsection{Experimental Setup}

As illustrated in Figure~\ref{fig:pipeline}, we evaluate models in a VQA format by querying,
\textit{``What is the name of the food, and which country is it most closely associated with?''}. In \MF and \MFB, the target food is placed on the left and distractor on the right. Thus, we specify \textit{``left food''} in place of \textit{``food''} in the prompt for \MF and \MFB. 
We measure the \textbf{food and country identification accuracy} of the target food.
Food identification accuracy is computed using similarity matching with a predefined threshold. Country identification accuracy is computed using exact string matching, while accounting for known variations in country names.\footnote{For food accuracy, qe use a weighted Jaccard character n-gram similarity (0.7 bigrams, 0.3 unigrams) with a threshold of 0.4. Two authors validated the evaluation method by manually reviewing 100 randomly sampled prediction–evaluation pairs from Gemini and InternVL-8B, confirming 95\% correctness for food-name scoring and 100\% for country scoring.} We further examine potential biases related to the position of the target item (left \emph{vs.} right) and its relative size. Both factors show negligible influence on model performance. (See Appendix~\ref{appendix:bias_ablations}.)

\vspace{0.5em}
\noindent\textbf{Models.} 
We evaluate 10 LVLMs, including 2 proprietary models (GPT-5 (\texttt{gpt-5-2025-08-07})~\citep{openai2025gpt5}, Gemini-2.5-Pro~\citep{comanici2025gemini}) and 8 open-source models.
We test the open-source models spanning parameter sizes from 8B to 72B: 
InternVL3 (8, 14, 38, and 78B)~\citep{zhu2025internvl3}, Ovis2.5-9B~\citep{lu2024ovis,lu2025ovis25technicalreport}, QwenVL3 series (8 and 32B)~\citep{Qwen2.5-VL}, and Molmo-72B~\citep{deitke2024molmo}.

\subsection{Overall Prediction Accuracy on \task}

Figure~\ref{fig:main_models_acc} presents the accuracy of all evaluated models on our \task.

\paragraph{Challenges in Understanding Culturally Mixed Images.} 
Across subtasks, the decrease in performance from \SF to \MF, \SFB, and \MFB across all models indicates that models experience greater difficulty in recognizing target food items under culture-mixing conditions. We observe a trend across models: \( \text{\SF} \gtrsim \text{\MF} > \text{\MFB} \gtrsim \text{\SFB} \).
Compared with food-item distractors (\MF), the background distractors (\SFB) result in a 13\% larger drop in country accuracy and a 7\% larger drop in food name accuracy on average. Street and landmark backgrounds show similar effects (see Figure~\ref{fig:app_country_heatmap} in Appendix~\ref{appendix:results}).

\vspace{0.5em}
\noindent\textbf{Performance Gap between Proprietary and Open Models.}
Both Gemini and GPT substantially outperform all open models across our subtasks, demonstrating stronger multimodal reasoning and cultural understanding. Among the open models, OVIS2.5-9B performs most competitively despite its relatively small parameter size, followed by InternVL3-78B, QwenVL3-32B, and QwenVL3-8B.


\subsection{Country Prediction Patterns}
Figure~\ref{fig:main_models_heatmap} compares the country prediction distributions across subtasks. 

\paragraph{Skewed Predictions Toward High-Resource Countries in \SF.}
In \SF, models relatively predict correct country labels, as indicated by the prominent blue diagonal trend. Accuracy is particularly high for countries such as India, Korea, Japan, and China among Asian regions, Italy and Poland among European regions, and the United States.
When misidentifying the country in the \SF subtask, models tend to confuse countries within the same continent (as seen in the Asian and European regions of the \SF heatmap). Additionally, predictions are often biased toward high-resource (WEIRD~\cite{henrich2010weirdest}) countries. Specifically, African and Asian countries are frequently misclassified as India or China, while European and North American countries are often mislabeled as the United States.
\paragraph{Prediction Shifts under Culturally Mixed Settings.}
Under the culturally mixed settings (\MF, \SFB, \MFB), accuracy declines, and model predictions are distributed more broadly across regions, deviating from the pattern observed in \SF.  Notably, in \SFB, predictions exhibit strong clustering around certain countries such as Mexico or Japan, in contrast to the high-resource region bias observed in \SF. We further analyze these patterns in depth in \S~\ref{sec:discussion}.


\section{Discussion}
\label{sec:discussion}
We examine how cultural distractors shift model predictions (§\ref{sec:discussion-effects}), how cultural distance affect these shifts (§\ref{sec:discussion-distance}), how they compare with culturally agnostic distractors (§\ref{sec:discussion-type}), and whether single-food cultural awareness correlates with robustness in culture mixing contexts (§\ref{sec:discussion-accuracy}).

We utilize three complementary metrics:
1) \textbf{accuracy},
2) \textbf{entropy} for prediction confidence and consistency, and
3) \textbf{label shift (\%)}, the proportion of predictions that differ from the \SF baseline in mixed settings, indicating how strongly distractors alter model outputs.



\subsection{Effects of Cultural Distractors on LVLMs}
\label{sec:discussion-effects}

\begin{figure}[t]
    \centering
    \begin{subfigure}[t]{0.51\columnwidth}
        \centering
        \includegraphics[width=\columnwidth]{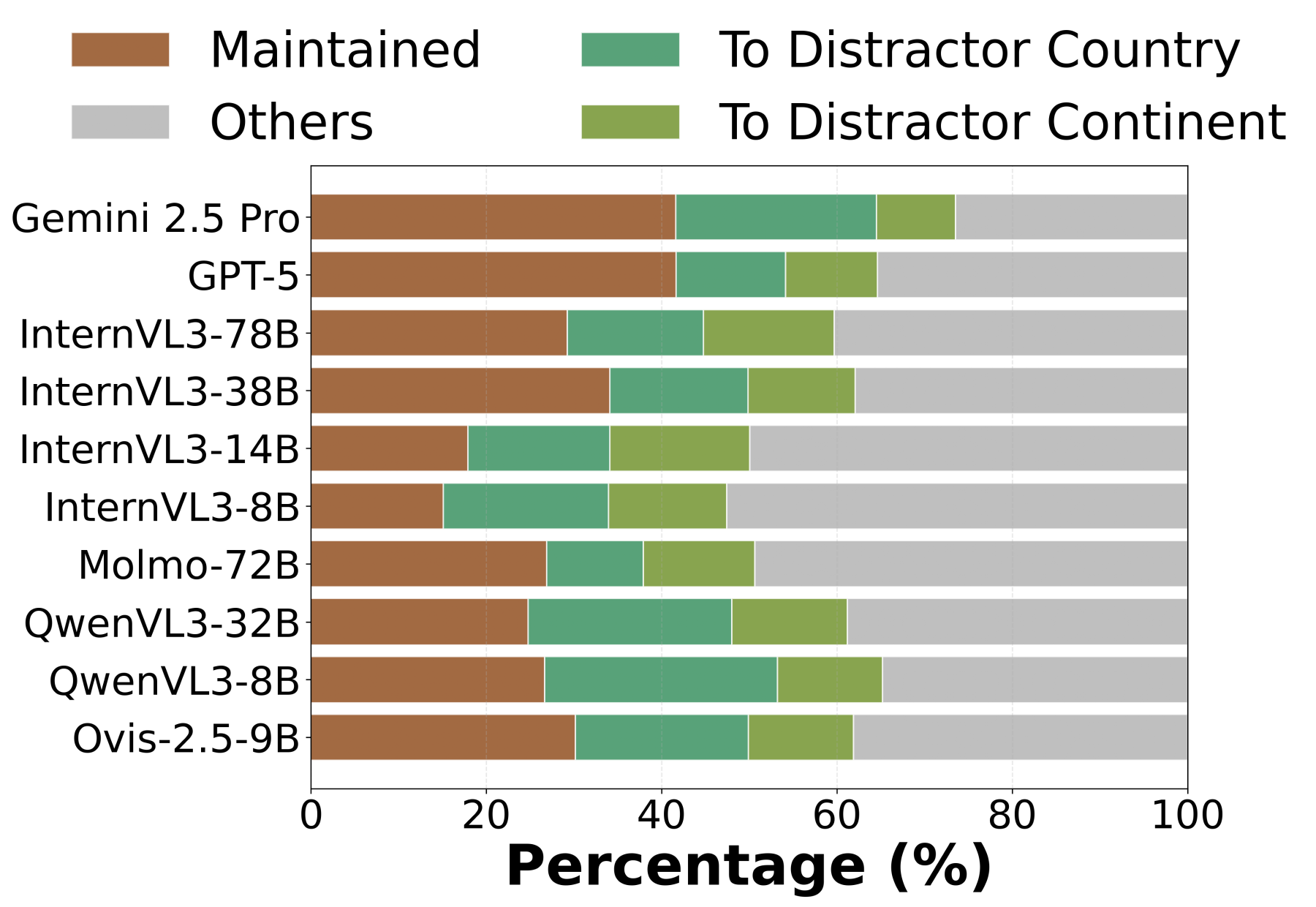}
        \caption{Prediction shifts across models}
        \label{fig:pred_label_shift_model}
    \end{subfigure}
    \begin{subfigure}[t]{0.48\columnwidth}
        \centering
        \includegraphics[width=\columnwidth]{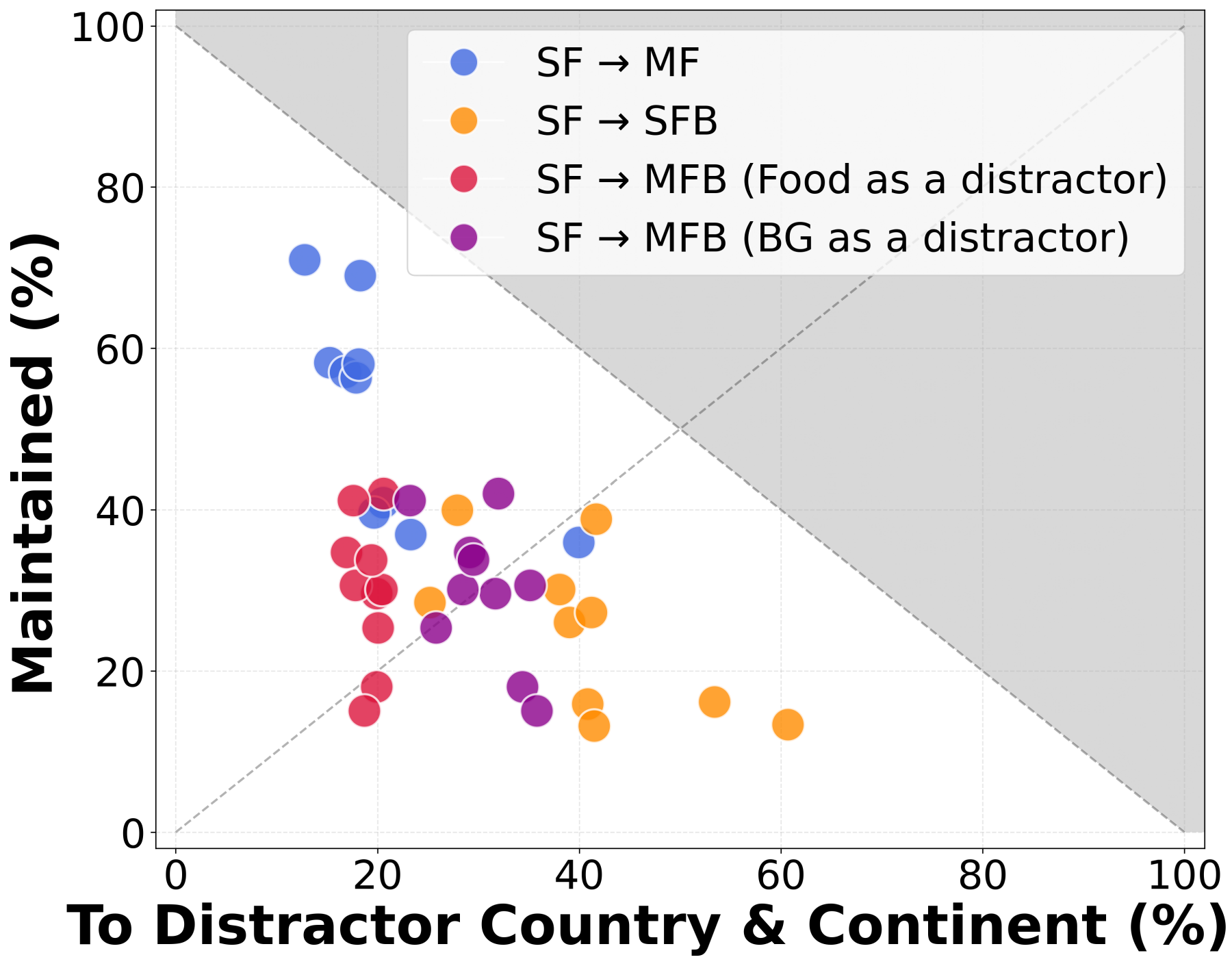}
        \caption{Prediction shifts across subtasks}
        \label{fig:pred_label_shift_subtask}
    \end{subfigure}
    \caption{\textbf{Effect of cultural distractors on country prediction label shifts.}}
    \label{fig:pred_label_shift_pattern}
\end{figure}

\paragraph{Distractors pull prediction shift toward their culture.} Figure~\ref{fig:pred_label_shift_model} shows how model predictions change when a distractor is added.
Across all models, an average 15\% of the predictions shift directly to the distractor's country, and an additional 12\% shift to a country in the same continent, indicating clear directional influence from culturally related distractors.
While Gemini achieves higher accuracy than GPT-5, its country predictions are more susceptible to distractor influence, exhibiting larger label shifts toward the distractor's country. Among open models, QwenVL3-8B and QwenVL3-32B show the strongest susceptibility, with the highest deviations toward the distractor's country (26.6\%, 23.3\%) under culture mixing settings.

\paragraph{Background Distractors Exert Stronger Influence Than Food Items.}
Figure~\ref{fig:pred_label_shift_subtask} compares distractor influence across subtasks by plotting the proportion of predictions that maintain \SF labels against those that shift toward the distractor.
Models appearing in the upper-left region of the plot are more resistant to distractors, whereas those in the lower-right region are highly affected. 
\MF shows high retention (40-80\%) with low shift ($\sim$20\%), showing models' robustness to food-item distractors. In contrast, \SFB yields lower retention (20-40\%) and higher shifts ($\sim$40\%), confirming that background cues provide stronger cultural signals than food items alone. \MFB shows performance between that of \SFB and \MF. Specifically, when food and background cues align, they reinforce predictions, but when they conflict, they increase shifts. This balance places \MFB between the food-only and background-only conditions.

\subsection{Effect of Elements' Cultural Distance on Prediction}
\label{sec:discussion-distance}
\begin{figure}[t]
    \centering
    \includegraphics[width=\columnwidth]{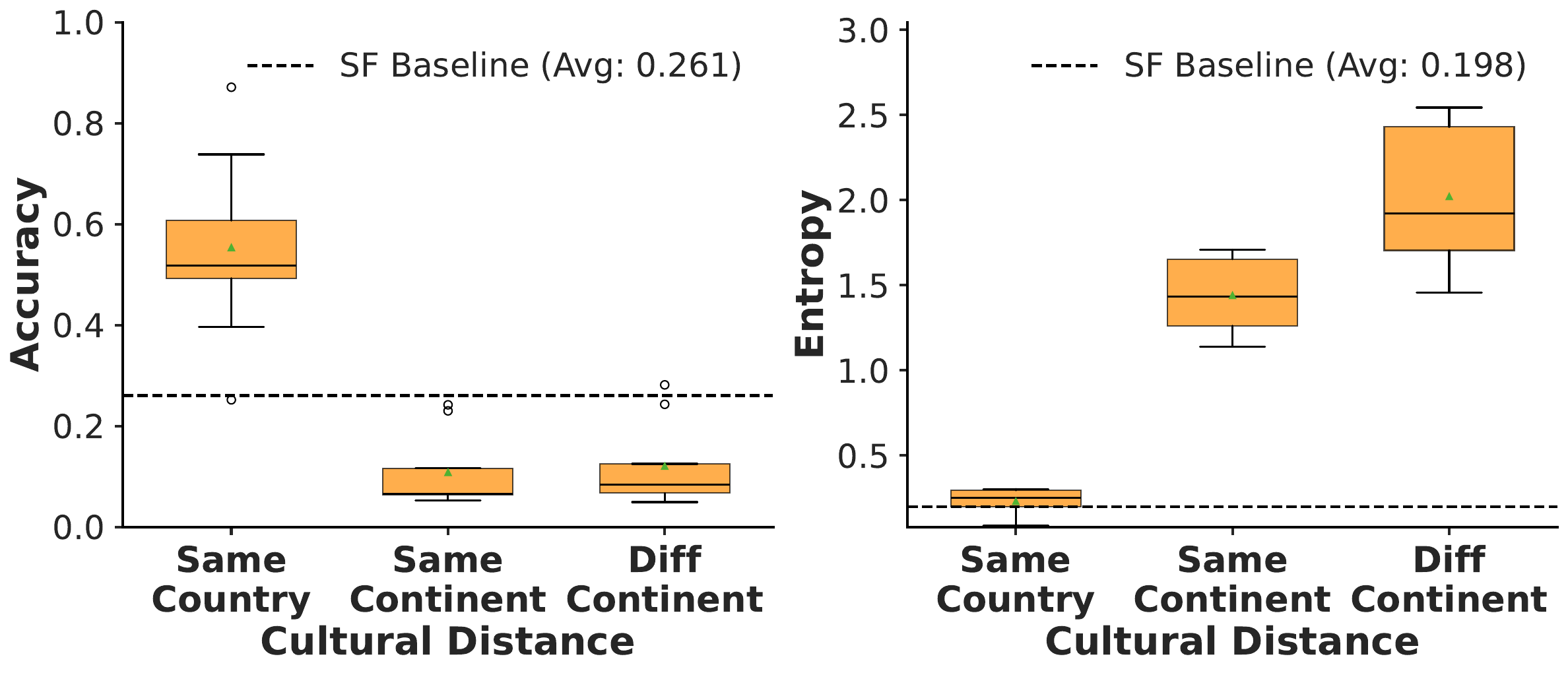}
    \caption{\textbf{Effect of cultural distance between target food and background distractor in \SFB on country identification accuracy and prediction entropy.}}
    \label{fig:cult_dist_acc_main}
\end{figure}

We examine how \textit{cultural distance} (whether the target and distractor originate from the same country, same continent, or different continents) affects model accuracy. 
As shown in Figure~\ref{fig:cult_dist_acc_main}, models achieve the highest accuracy and lowest entropy when the target and distractor come from the same country, which suggests that culturally consistent cues enhance recognition by providing coherent context. In contrast, accuracy is lowest and entropy is highest when the target and distractor originate from different continents.
Similar patterns appear in \SF, \MFB~settings and food name identification task, as detailed in Figure~\ref{fig:cultural_distance_analysis} in Appendix~\ref{appendix:results}, which shows consistent monotonic improvement with decreasing cultural distance.

\subsection{Comparison Between Cultural and Culturally-Agnostic Distractors}
\label{sec:discussion-type}
\begin{figure}[t]
    \centering
    \includegraphics[width=\columnwidth]{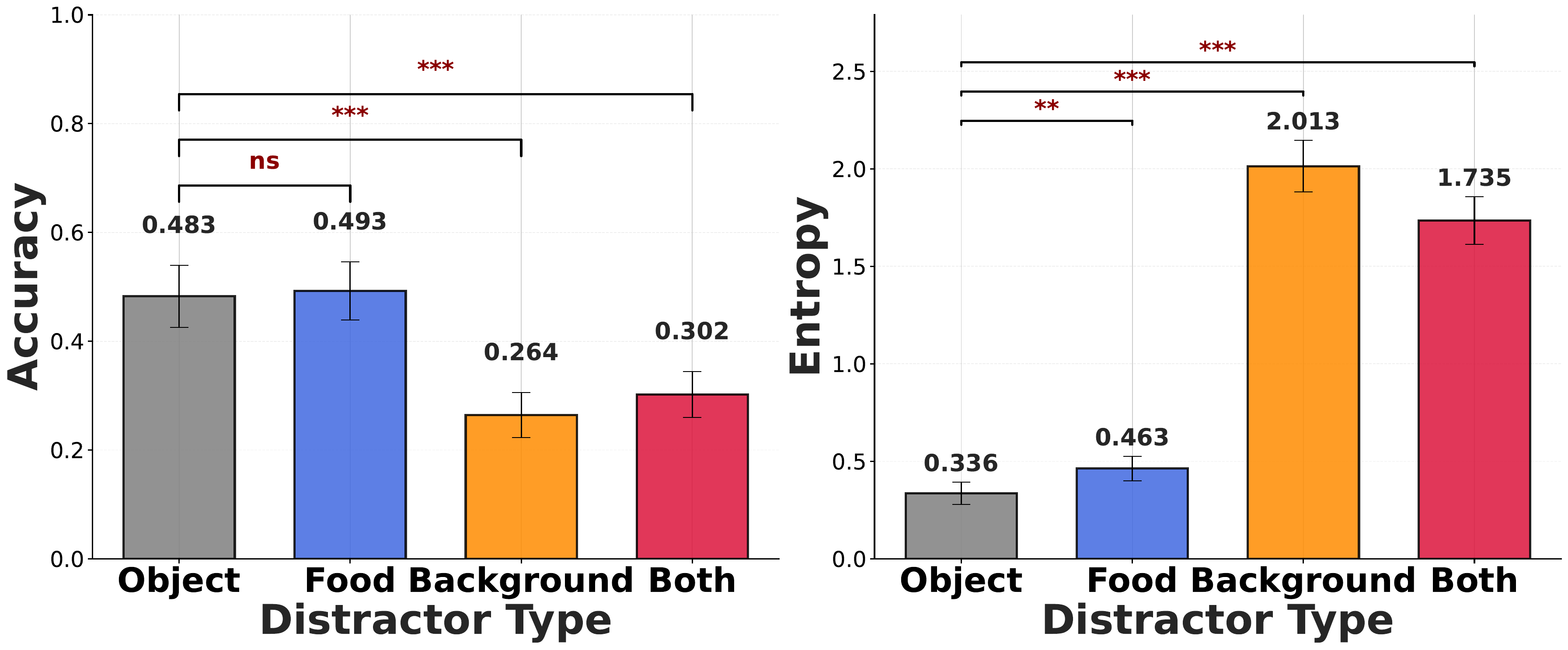}
    \caption{\textbf{Comparison of distractor types on country identification accuracy and prediction label entropy (Gemini).} (ns: not significant, **:p$<$0.01, ***: p$<$0.001)}
    \label{fig:distractor_compare}
\end{figure}

To verify that these effects stem from cultural information rather than general distractor complexity, we repeat our experiment using four culturally-agnostic objects (apple, car, scissors, and teddy bear) as distractors. As shown in Figure~\ref{fig:distractor_compare}, cultural distractors generally yield similar or lower accuracy and higher entropy than agnostic ones, confirming that cultural signals drive the observed shifts, rather than visual content alone.
Although distractors sometimes yield slightly higher accuracy than object distractors, this is primarily due to the distractors from the same country enhancing prediction (\S~\ref{sec:discussion-distance}).
Consistent with \S~\ref{sec:discussion-effects}, background distractors have stronger effects than food distractors, shown by the largest accuracy drops and highest entropy increases.
\subsection{Relationship Between Cultural Awareness of Single Foods and Multi-Foods}
\label{sec:discussion-accuracy}

\begin{figure}[t]
    \centering
    \includegraphics[width=\columnwidth]{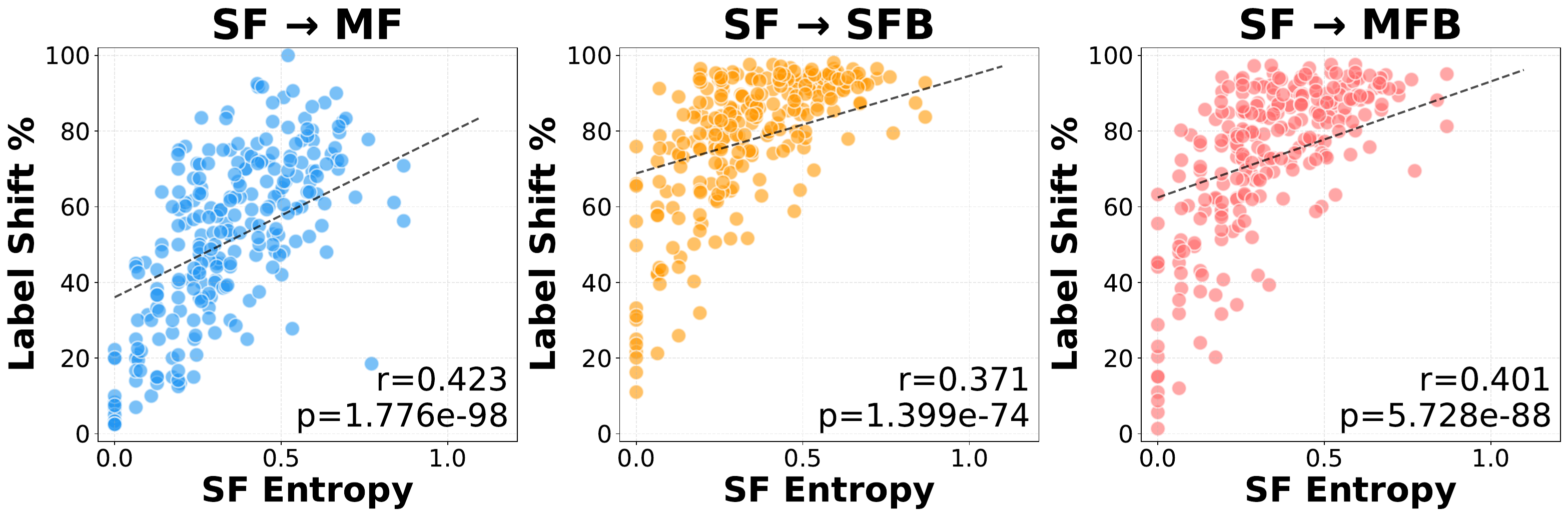}
    \caption{\textbf{Relationship between model confidence in the \SF setting and the label shift ratio under cultural mixing (\MF–\MFB) settings.} Confidence is measured by entropy (lower entropy = higher confidence) Each point represents a Country Identification Accuracy of each food item.}\label{fig:scatter_confidence_vs_label_shift}
\end{figure}

Finally, we examine whether a model’s understanding of individual foods predicts robustness in mixed-cultural settings. Figure~\ref{fig:scatter_confidence_vs_label_shift} relates \SF entropy (a proxy for prediction confidence) to label shift under culture mixing.
Lower confidence generally corresponds to larger shifts, indicating that weaker item-level understanding leaves models more vulnerable to distractors.
However, low entropy does not guarantee robustness in culturally mixed contexts, as several low-entropy foods still exhibit high shifts. These cases imply that distractor resistance depends not only on target knowledge but also on the relative strength of competing visual cues in the image.

\section{\task-Real: Real-World Culturally Mixed Food Dataset}
\subsection{Dataset Collection}
To assess whether the behaviors observed in our synthetic datasets also arise in natural images in the real world, we collect a real-world culture mixing dataset (for \MF setting). Images were gathered with web-based image searches using systematically generated combinations of culturally specific keywords (\eg, ``Korean and Italian food on a table''), and a survey within our institute, where participants submitted photos with explicit consent.
We collect 100 \MF images, consisting of 50 same-culture and 50 cross-culture food combinations spanning 10 countries. By cropping individual food regions from these \MF~images, we extract 219 single-food (\SF) images, enabling a direct comparison between isolated and culturally mixed contexts.

\subsection{Experiment and Results}
\begin{figure}[t]
    \centering
    \includegraphics[width=\columnwidth]{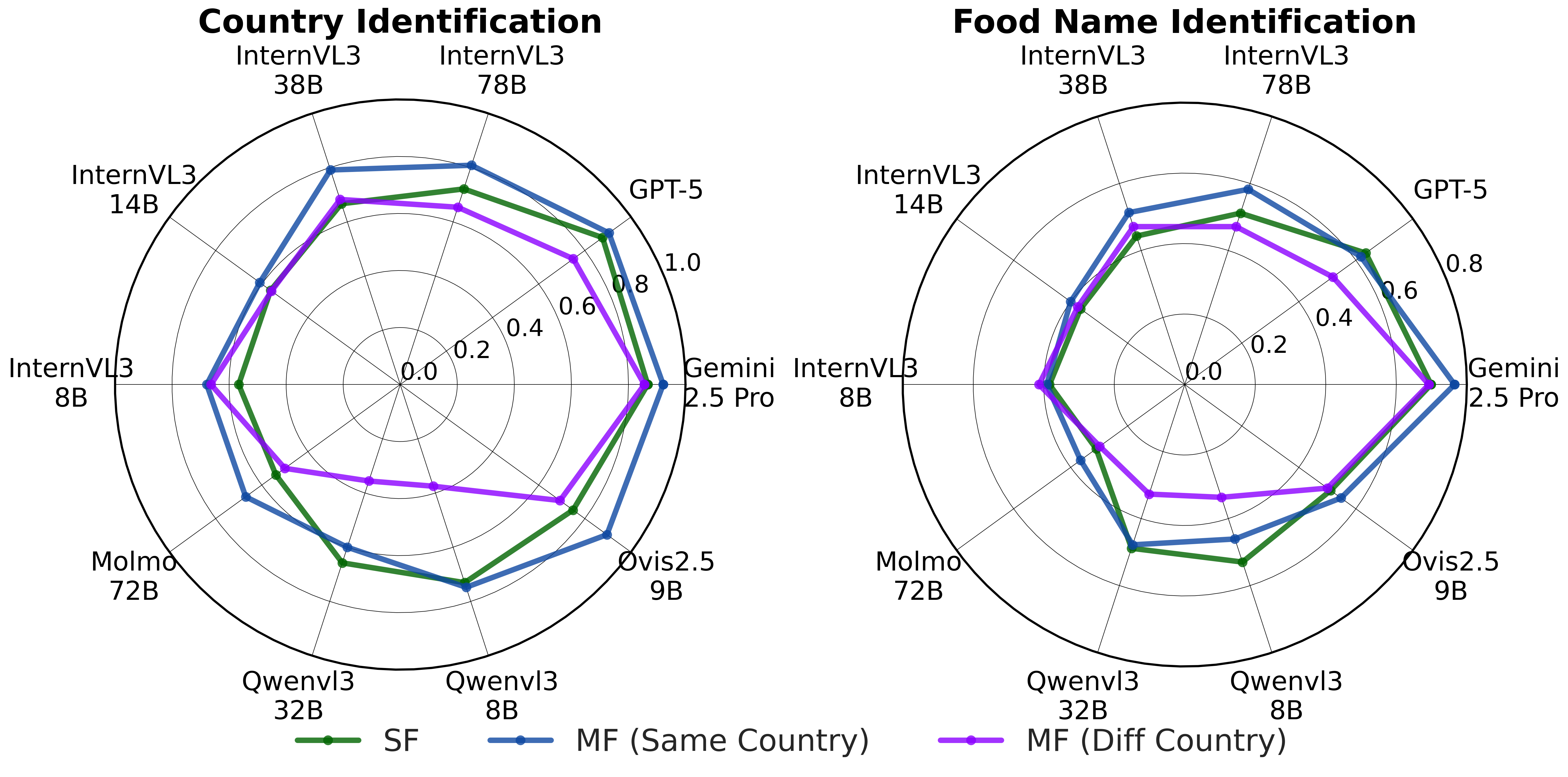}
    \caption{\textbf{Overall prediction accuracy in \task-real across models.}}
    \label{fig:real_models_acc}
\end{figure}

We evaluate the same set of models described in \S~\ref{sec:experiments}. Because food items in real-world images are not always symmetrically positioned (\eg, side by side), we mark the target food with a red bounding box and prompt the model to identify the food within the bounding box.

As shown in Figure~\ref{fig:real_models_acc}, the results reveal a trend of \MF (same) $>$ \SF $>$ \MF (diff), indicating that the cultural relationship between co-occurring elements affects the model performance in the real world as well.
This confirms that the degradation of cultural understanding observed in our synthetic \task settings also manifests in real-world scenarios, underscoring the persistent challenge of LVLMs.

\section{Improving Understanding of Culture Mixing: An Exploratory Study}
As observed in \S~\ref{sec:experiments}-\ref{sec:discussion}, existing LVLMs struggle with culturally mixed scenarios, yet understanding such contexts is critical for building culturally aware models for real-world use. This leaves us with an important open question: \textit{How can we build models that are capable of handling culture mixing scenarios?} Building on the insights from our analysis, we conduct several exploratory approaches as a starting point to address this challenge, examining both training-free and training-based methods.
We use two open source models, Ovis2.5-9B and InternVL3-8B, in our experiment.
\vspace{-0.5em}
\paragraph{Training and Test Datasets.} 
From the entire set of datasets included in \task, we create train-test splits as follows: First, \SF images are divided with a 7:3 train-test ratio, ensuring balanced country distribution across both sets. \MF, \SFB, and \MFB datasets are then split correspondingly based on the inclusion of their associated \SF images. For the training dataset used in our training-based approach, we randomly sample one-third of instances from each food item across all four subsets (\SF, \MF, \SFB, and \MFB), resulting in 5K training images, to reduce the overall training cost. For evaluation, we use the complete test set of 7K images to assess all three mitigation methods.

\subsection{Approaches for Improvement}
Through evaluation on \task, we reveal that: 1) even when LVLMs demonstrate strong knowledge or confidence about certain items in isolation, they fail when these items appear in culturally mixed contexts, and 2) this failure stems heavily from their reliance on co-occurring cultural items and backgrounds. These findings indicate that understanding culture mixing does not necessarily align with cultural awareness of individual items, requiring approaches that go beyond single-item recognition.

\begin{enumerate}
    \item \textbf{Direct Prompting Engineering (\textbf{$Prompt_{Direct}$})}
    We explicitly instruct the model to focus on the target item rather than the background or other elements by adding direct guidance to the prompt. This approach aims to mitigate the model's reliance on surrounding cultural cues identified in our analysis.

    \item \textbf{Chain of Thought (CoT) Prompting (\textbf{$Prompt_{CoT}$})}
    We explore Chain-of-Thought (CoT) prompting because culture mixing often requires models to integrate multiple, partially informative visual cues~\citep{wei2022chain}. Such multi-cue integration amounts to a form of visual–contextual reasoning, and CoT has been shown to improve LLM performance on various tasks.
    \item \textbf{Supervised Finetuning ($SFT$)}
     We fine-tune models on culturally mixed images to help them learn to ignore cultural distractors. During training, each food item is presented in the following order: \SF - \MF - \SFB - \MFB.
This progression exposes the model to increasingly complex cultural mixtures while encouraging consistent predictions across these varying contexts.
\end{enumerate}

\begin{table}[t]
\centering
\small
\caption{\textbf{Performance of Ovis-2.5-9B and InternVL3-8B across mitigation methods.} 
Lower is better for Entropy, higher is better for Accuracy. 
\underline{Underline} indicates statistically significant improvement; 
\cellcolor{gray!15}{gray boxes} indicate results better than the Base setting.}
\label{tab:mitigation}
\resizebox{0.95\columnwidth}{!}{%
\begin{tabular}{l|ccc|cccc}
\toprule
 & \multicolumn{3}{c|}{\textbf{Entropy (↓)}} & \multicolumn{4}{c}{\textbf{Accuracy (↑, \%)}} \\
\cmidrule(lr){2-8}
\textit{Model (Setting)} & MF & SFB & MFB & SF & MF & SFB & MFB \\ 
\midrule
\textbf{\texttt{Ovis2.5}} &&&&&&& \\
\hspace{1em}\textbf{$Base$} 
 & 1.28 & 3.34 & 3.07 
 & 13.89 & 10.74 & 5.65 & 6.14 \\

\hspace{1em}\textbf{$Prompt_{Direct}$}   
 & \cellcolor{gray!15}\textbf{0.97} & \cellcolor{gray!15}3.31 & \cellcolor{gray!15}2.99 
 & \cellcolor{gray!15}\textbf{16.67} & \cellcolor{gray!15}\textbf{14.07} & \cellcolor{gray!15}6.00 & \cellcolor{gray!15}6.07 \\

\hspace{1em}\textbf{$Prompt_{CoT}$}   
 & \cellcolor{gray!15}1.24 & 3.60 & 3.21 
 & \cellcolor{gray!15}15.28 & \cellcolor{gray!15}14.81 & \cellcolor{gray!15}6.62 & \cellcolor{gray!15}6.73 \\

\hspace{1em}\textbf{$SFT$} 
 & \cellcolor{gray!15}\underline{1.02} & \cellcolor{gray!15}\textbf{\underline{2.52}} & \cellcolor{gray!15}\textbf{\underline{2.36}} 
 & \cellcolor{gray!15}15.28 & \cellcolor{gray!15}11.85 & \cellcolor{gray!15}\textbf{\underline{8.59}} & \cellcolor{gray!15}\textbf{\underline{8.95}} \\ 
\midrule

\textbf{\texttt{InternVL3}} &&&&&&& \\
\hspace{1em}\textbf{$Base$} 
 & 1.16 & 3.77 & 3.43 
 & \textbf{11.11} & 5.19 & 2.14 & 3.33 \\

\hspace{1em}\textbf{$Prompt_{Direct}$}  
 & \cellcolor{gray!15}\textbf{1.12} & \cellcolor{gray!15}3.77 & \cellcolor{gray!15}3.36 
 & 9.72 & \cellcolor{gray!15}\textbf{8.52} & \cellcolor{gray!15}\textbf{2.62} & \cellcolor{gray!15}{3.47} \\

\hspace{1em}\textbf{$Prompt_{CoT}$}  
 & 1.54 & 3.97 & 3.65 
 & 9.72 & \cellcolor{gray!15}7.78 & \cellcolor{gray!15}2.65 & 2.74 \\

\hspace{1em}\textbf{$SFT$} 
 & \cellcolor{gray!15}\underline{1.13} & \cellcolor{gray!15}\textbf{\underline{2.76}} & \cellcolor{gray!15}\textbf{\underline{2.45}} 
 & 9.72 & \cellcolor{gray!15}\underline{8.15} & \cellcolor{gray!15}\textbf{\underline{4.16}} & \cellcolor{gray!15}\underline{\textbf{5.14}} \\
\bottomrule
\end{tabular}%
}
\end{table}

\begin{table*}[t]
\centering
\caption{\textbf{Comparison of cultural benchmark datasets.} This table summarizes existing datasets in terms of cultural content, task type, image modality, geographic and linguistic coverage, and the presence of culture mixing. 
Our benchmark dataset is the first to explicitly include \textbf{cultural mixing}, covering diverse foods and backgrounds across 30 countries in English, and supporting both real and synthetic images.}
\label{tab:realwork}
\resizebox{\linewidth}{!}{%
\begin{tabular}{@{} >{\ttfamily}llllllll}
\toprule
& \textbf{Cultural Element} & \textbf{Evaluation Type} & \textbf{Task Type} & \textbf{Image Type} & \textbf{Countries} & \textbf{Languages} & \textbf{Culture Mixing} \\
\midrule
Bhatia et al. (2024)~\citep{bhatia-etal-2024-local} & Object & Retrieval & Visual grounding & Real & 50 countries & English & No \\
Yin et al. (2021)~\citep{yin-etal-2021-broaden} & Scene & VQA & Commonsense reasoning & Real & 4 regions & English & No \\
Vayani et al. (2025)~\citep{Vayani_2025_CVPR} & Food, Scene, Object & VQA, Captioning & Cultural knowledge & Real & 73 countries & 100 languages & No \\
Romero et al. (2024)~\citep{NEURIPS2024_1568882b} & Scene, Object & VQA & Cultural knowledge & Real & 30 countries & 31 languages & No \\
Nayak et al. (2024)~\citep{nayak-etal-2024-benchmarking} & Object, Scene & VQA & Cultural knowledge & Real & 11 countries & English & No \\
Nikandrou et al. (2025)~\citep{nikandrou-etal-2025-crope} & Object & VQA & Contextual adaptation & Real / Hybrid & 5 Countries & 5 languages & No \\
Zhou et al. (2025)~\citep{zhou-etal-2025-mapo} & Food (text only) & Probing & Cultural knowledge & Mostly Text & 14 countries & 6 Languages & No \\
Kim et al. (2025)~\citep{kim-etal-2025-tom} & Ethnicity, Background & VQA & Cultural bias & Hybrid & 5 countries & English & Yes \\
\rowcolor{gray!15} \task & Food, Background & VQA & Cultural knowledge & Both & 30 countries & English & \textbf{Yes} \\
\bottomrule
\end{tabular}%
}
\end{table*}

\subsection{Results}

Table~\ref{tab:mitigation} compares the entropy and accuracy of three mitigation methods against the base model.
\vspace{-1mm}
\paragraph{Direct prompting and SFT consistently enhance LVLMs' performance.} Overall, Direct prompting and SFT consistently improve performance across most tasks, though only SFT achieves \textit{statistically significant} gains (paired T-test, \textit{$p<.01$}). For \MFB and \SFB tasks, SFT remarkably reduces entropy by a substantial margin while also increasing accuracy, notably leading to more robust performance. In the \MF setting, while SFT decreases entropy, Direct prompting demonstrates marginally better effectiveness for robustness. These results suggest that training-free strategies may be sufficient for simple mixing scenarios, but as contexts become more complex, particularly when background elements are involved, training-based approaches like SFT become necessary to adequately equip models with the capability to recognize culturally mixed situations.
\vspace{-0.5em}
\paragraph{CoT prompting does not always help.} Our results show that CoT prompting sometimes improves accuracy on \task, but fails to enhance and occasionally even degrades label consistency. To understand this discrepancy, we analyze the cases where CoT introduces new errors, and find that it often magnifies the model’s reliance on background cues, ultimately leading to incorrect predictions.
This finding aligns with recent research on LLM reasoning in cultural and social tasks~\cite{kim2025lovers}, which demonstrates that CoT can over-amplify the influence of misleading cues when models need to aggregate information from multiple, potentially conflicting sources/contexts.

As our analysis reveals, while prompting shows promise, its limitations highlight the need for training approaches. Our preliminary results indicate that supervised fine-tuning on culturally mixed scenarios yields meaningful improvements. We call for research into training objectives designed for culture mixing, essential for LVLMs operating in culturally diverse contexts.

\section{Related Work}
\label{sec:rel-work}
\subsection{Cultural Awareness of LVLMs}
The growing recognition that cultural differences impact visual understanding \citep{Ye_2025_CVPR, berger-ponti-2025-cross} has spurred significant interest in the cultural awareness of vision–language models (VLMs) \citep{10.1162/COLI.a.14}.
Consequently, researchers have investigated cultural biases across a wide range of tasks, including image captioning \citep{mohamed-etal-2024-culture, 10924504}, text-to-image generation \citep{kannen2024beyond, bayramli-etal-2025-diffusion}, retrieval and visual grounding \citep{bhatia-etal-2024-local}, and safety \citep{bui-etal-2025-multi3hate, yerukola-etal-2025-mind}. 
Common evaluation approaches include the Visual Question Answering (VQA) setting \citep{yin-etal-2021-broaden, Vayani_2025_CVPR, NEURIPS2024_1568882b, nayak-etal-2024-benchmarking, nikandrou-etal-2025-crope} and the use of culturally specific datasets, for instance those featuring food from various countries \citep{winata-etal-2025-worldcuisines, zhou-etal-2025-mapo}.
However, these datasets generally contain images rooted in a single cultural context, making it difficult to evaluate how models interpret cultural ambiguity or blended settings.
In particular, \citet{kim-etal-2025-tom} introduce a framework that replaces individuals in real images with people of different ethnic backgrounds to reveal ethnicity-driven biases in recognizing cultural elements.
While their work explores the intersection of ethnic background and cultural artifacts, it focuses on isolated elements rather than scenarios where multiple cultural artifacts (\eg, food and clothing) coexist within the same visual scene.

As shwon in Table~\ref{tab:realwork},
while existing datasets cover various combinations of objects, scenes, and food items across different countries and languages, none explicitly target \textbf{culture mixing} in visual content.
In contrast, \task{} systematically constructs images where multiple cultural elements co-occur, encompassing diverse foods and backgrounds across 30 countries and supporting both real and synthetic images.
To the best of our knowledge, this is the first work to leverage such culture-mixing scenarios to reveal how LVLMs behave under cross-cultural ambiguity and to provide a benchmark for evaluating and improving their cross-cultural reasoning and generalization.



\subsection{Image Fusion and Composition}
Image fusion, composition, and blending have been extensively studied for applications including object detection, image generation, and data augmentation.
Early research focused on integrating information from different sensor modalities to enrich visual data \citep{electronics12010097,liu2022target}.
Image composition methods have facilitated the automated creation of synthetic data and augmentation strategies \citep{Hao_2023_WACV, 10.1609/aaai.v38i3.28025}.
These approaches not only enhance model generalization and pre-training efficiency but also provide scalable alternatives to manual annotation.
The advent of generative models has enabled the synthesis of multiple concepts and the transfer of visual styles or semantics with high realism \citep{liu2022compositional, kumari2022customdiffusion, chefer2023attendandexcite, liew2022magicmix}.
While existing work has primarily focused on enhancing model capabilities, our research pivots to use these techniques for evaluation.
We repurpose image composition to create challenging visual scenarios by combining images with diverse cultural elements to assess the cultural awareness of vision-language models.


\section{Conclusion}
This study presents a systematic evaluation of LVLMs to investigate their capabilities in culturally entangled visual contexts. By introducing a large-scale benchmark of global food and scene images, we reveal that while current LVLMs generally perform well when presented with a single food image (\SF), the performance decreases with the existence of cultural distractors (\MF, \MFB, and \SFB). Our results suggest that although these models regard new cultural components as distractors rather than integrating them as complementary cues, there is a need for interventions to improve cross-cultural understanding. To address this, we present training-free and training-based approaches on medium-sized open-source models to guide future works in developing mixed-cultural-aware LVLMs. We believe our findings and proposed approaches could provide a foundation for building LVLMs that reason more effectively across culturally diverse contexts.

\section*{Limitations and Future Work}
\label{sec:conclusion}

While our benchmark covers a wide range of global cuisines and scenes, it should be noted that there remains an underexplored cultural context, especially for underrepresented or low-resource regions. Also, our evaluation primarily focuses on food and scene recognition due to the challenges of collecting diverse, high-quality data (and the relatively lower controversy of these cultural elements). We leave for future work to explore other culturally rich domains, such as festivals, clothing, and daily-life interactions present in visual contexts. Finally, incorporating user-centered evaluations (other than ground-truth annotations) from diverse cultural backgrounds would provide richer insights into the practical effectiveness of LVLMs in real-world multicultural scenarios.

\section*{Acknowledgement}
We express our gratitude to Professor Diyi Yang for her insightful feedback, which shaped the direction of this research.

{
    \small
    \bibliographystyle{unsrtnat}
    \bibliography{main}
}
\clearpage

\appendix




\section{Pipeline Details of \task}
\label{appendix:pipeline-details}
\subsection{Country Lists and Statistics}
\label{appendix:stats}
\begin{figure}[ht]
    \centering
    \includegraphics[width=0.4\textwidth]{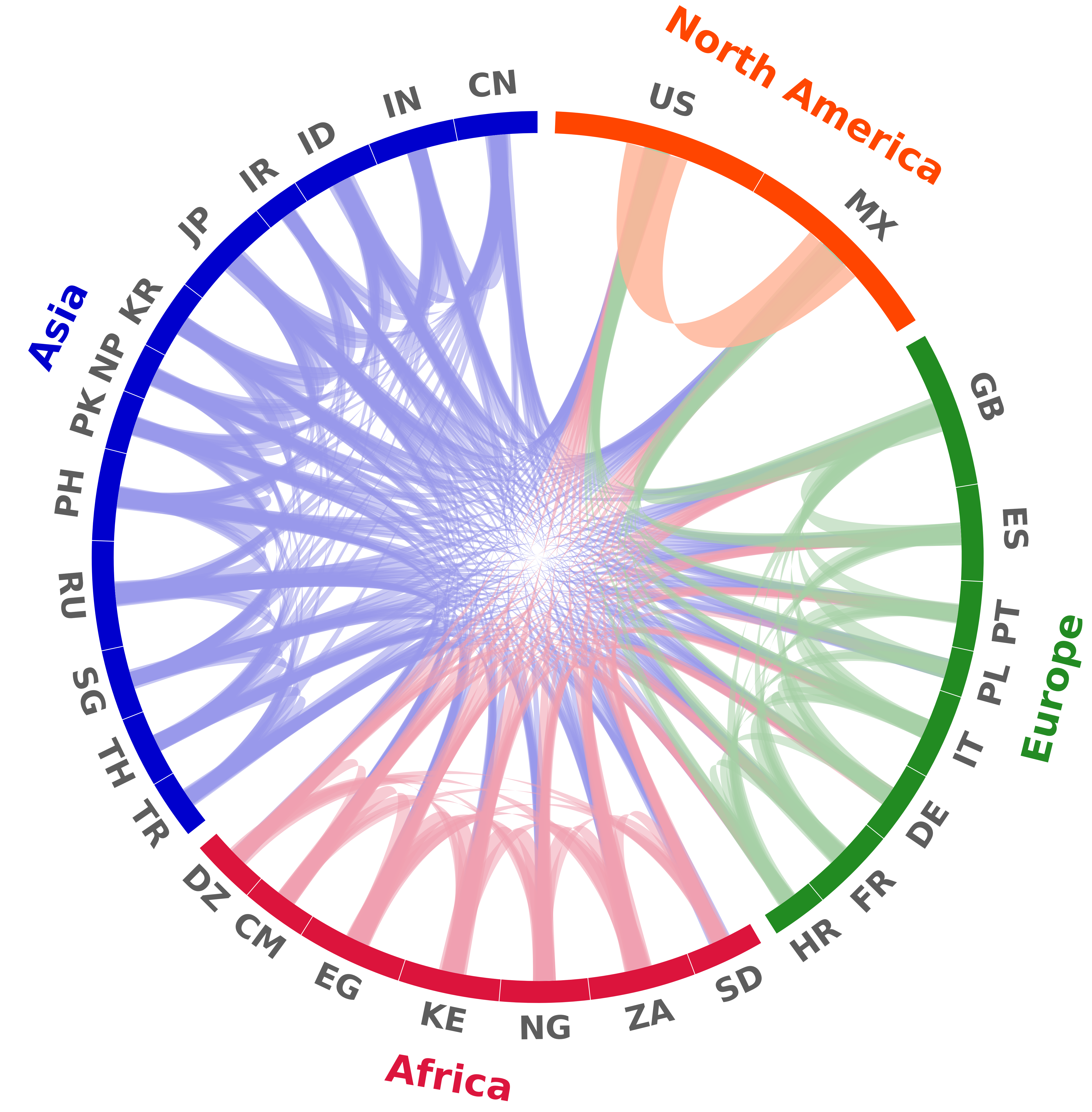}
    \caption{\textbf{Visualization of food combinations in \MF and \MFB.}} 
    \label{fig:food-comb}
\end{figure}

Table~\ref{tab:food_stats} summarizes, for each country, the number of food dishes and their names. Figure~\ref{fig:food-comb} visualizes the distribution of food combinations in \MF, showing that the combinations are diverse and well balanced.
\subsection{Human Validation and Statistics}
We incorporated a human-in-the-loop process throughout all stages of dataset construction. Human annotators validated the generated images, and those that did not meet the quality criteria were regenerated—either using the same model or a more advanced one—followed by another round of human validation. This iterative process was repeated until the final dataset was completed. The criteria used for validation and the statistics of filtered cases are as follows.

\paragraph{Generating Single Food (\SF) Images}
    \begin{enumerate}
        \item \textbf{Background removal}
        \begin{enumerate}
            \item If the image contains \textit{text, hands, people, or tableware}, these elements are removed using either a diffusion model or manual methods.
        \end{enumerate}
        \item \textbf{Human Validation--Comparison with the original image} (manually checked by the author using an in-house platform)
        \begin{enumerate}
            \item Criteria (Error Types)
            \begin{itemize}
                \item Image differs from the original $\rightarrow$ \textit{Regenerate}
                \item Food item is cropped $\rightarrow$ \textit{Regenerate}
                \item Tableware appears in the image $\rightarrow$ \textit{Regenerate}
                \item More than one food item appears $\rightarrow$ \textit{Regenerate}
                \item Text appears on the food $\rightarrow$ \textit{Manually hide the text}
                \item Reference food itself is inappropriate (e.g., food placed on a cauldron, making it unsuitable for table placement) $\rightarrow$ \textit{Replace food image}
                \item Other cases where the image appears unnatural $\rightarrow$ \textit{Replace food image}
            \end{itemize}
            \item Statistics
            \begin{itemize}
                \item Regeneration cases: 109 / 295
                \item Food image replacement cases: 12 / 295
               
                \item Images with text present: 1 / 295
            \end{itemize}
        \end{enumerate}
    \end{enumerate}

\paragraph{Generating Multiple Food (\MF) Images}
    \begin{enumerate}
        \item \textbf{Image Generation}
        \begin{itemize}
            \item Concatenate the two images generated in Step 2 to create a \MF image.
        \end{itemize}

        \item \textbf{Human Validation}
        \begin{enumerate}
            \item Criteria (Error Types)
            \begin{itemize}
                \item Is the food item cropped? $\rightarrow$ \textit{Regenerate}
                \item Are the two food items excessively unbalanced in size? $\rightarrow$ \textit{Regenerate}
            \end{itemize}
            \item {Statistics}
            \begin{itemize}
                \item Regeneration cases: 74 / 948
            \end{itemize}
        \end{enumerate}
    \end{enumerate}

\paragraph{Single Food with Background (\SFB) and Multi-Food with Background (\MFB) images}

    \begin{enumerate}
        \item \textbf{Background Image Selection}
        \begin{itemize}
            \item For each category (Street, Landmark), five background images were manually selected from different continents.
        \end{itemize}

        \item \textbf{Input Image Generation}
        \begin{itemize}
            \item Concatenate the \MF images beneath the selected background image.
        \end{itemize}

        \item \textbf{Editing}
        \begin{itemize}
            \item Example prompt (the first two versions were used when initial generations failed):
           
        \end{itemize}
        \item \textbf{Human Validation}
        \begin{enumerate}
            \item Criteria
            \begin{itemize}
                \item Background not generated (showing only the food item) $\rightarrow$ \textit{Regenerate}
                \item Food placed unnaturally (e.g., floating in the air or positioned upright as if it would spill) $\rightarrow$ \textit{Regenerate}
                \item Food or background differs from the original $\rightarrow$ \textit{Regenerate}
            \end{itemize}         
            \item {Statistics}
We repeated multiple rounds of refinement. After three attempts, most samples were collected as valid, with only a few remaining invalid. These remaining cases were not verified through the platform but were instead visually inspected and regenerated until satisfactory.\\
1st attempt: Valid/Total = 0.72 \\
2nd attempt: Valid/Total = 0.70\\
3rd attempt: Valid/Total = 0.88\\

        \end{enumerate}
    \end{enumerate}

\subsection{Prompts and Model Configurations for Synthetic Dataset Generation}
\label{appendix:generation_process}
\begin{figure}[t]
    \centering
    \includegraphics[width=0.8\columnwidth]{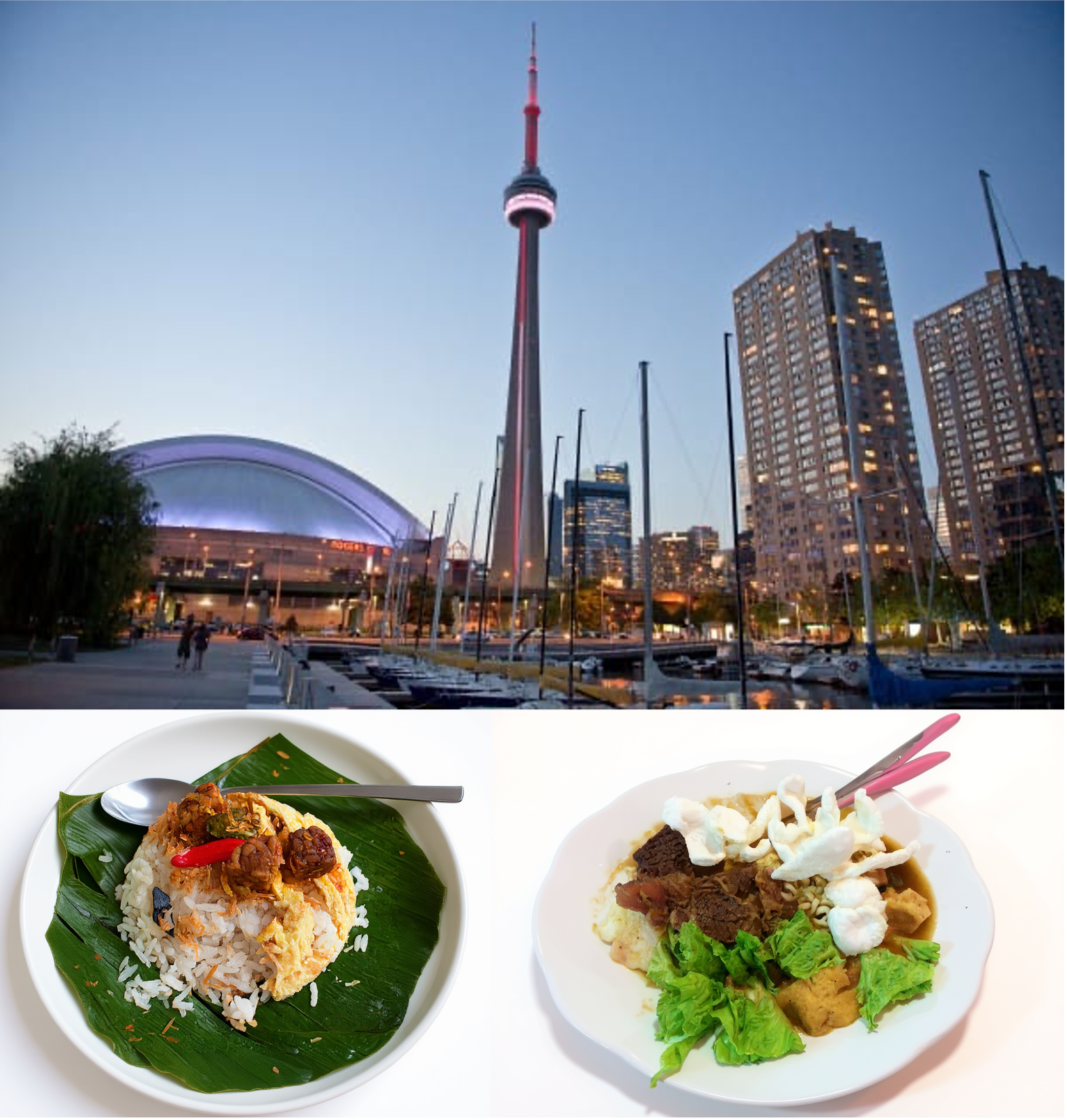}
    \caption{\textbf{An example of input Images for diffusion model based image editing.}}
    \label{fig:input}
\end{figure}

We provide details on how the image editing models were used to generate \SF, \MF, \SFB, and \MFB images. The specific model names and configurations are summarized in Table~\ref{tab:model-config}, and the corresponding prompts are provided below. Note that we used minor prompt variants (\eg, reordering sentences, substituting synonymous verbs) to regenerate images when the initial output failed to satisfy the required criteria of the human annotators. An example of input Images for diffusion model is shown in Figure~\ref{fig:input}.

\begin{figure*}[ht]
    \centering
    \includegraphics[width=0.65\linewidth]{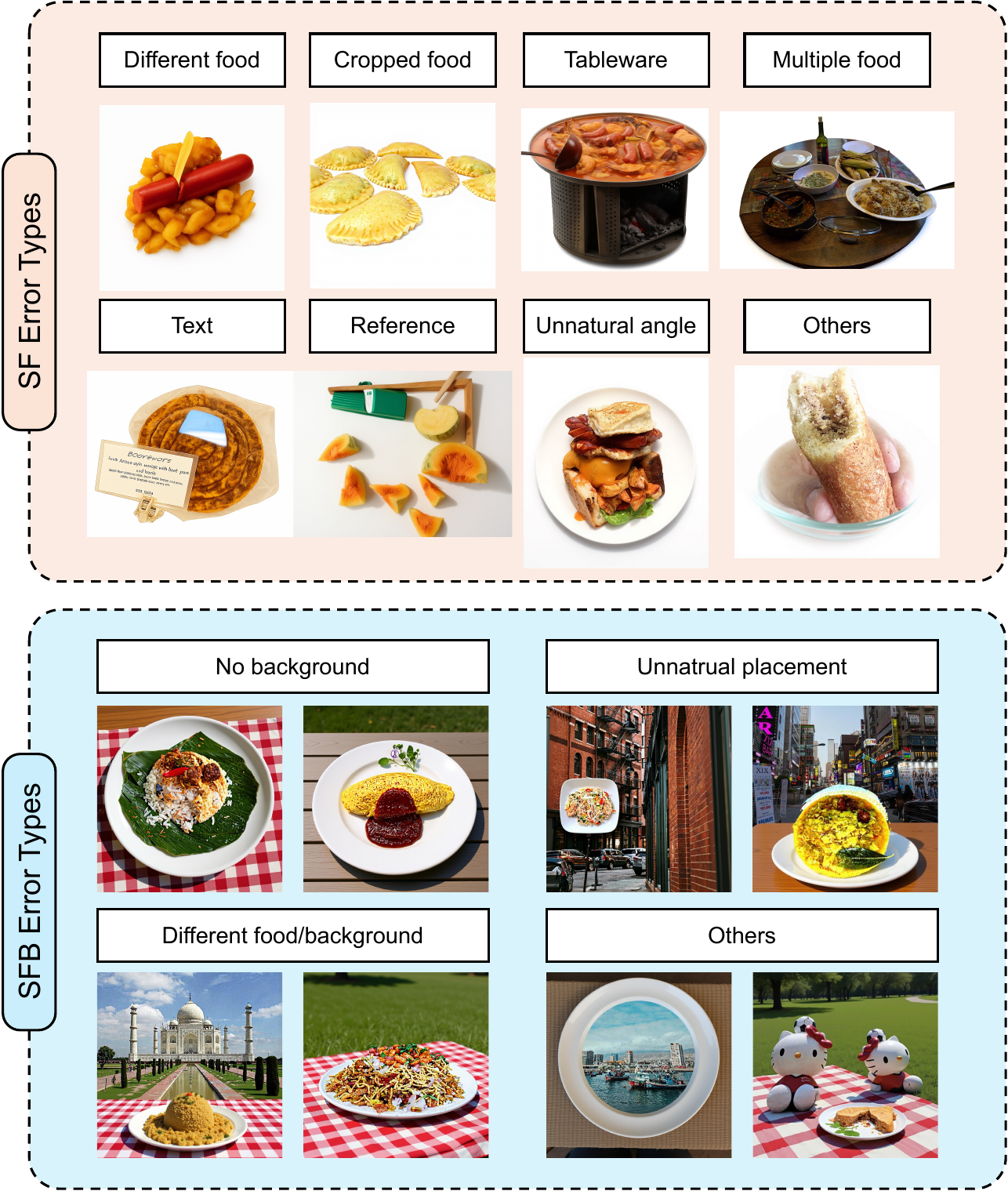}
    \caption{\textbf{Sample \SF and \SFB error images}. These images did not meet the human evaluation criteria and illustrate common failure patterns observed during the filtering process.}
    \label{fig:app_qual_error}
\end{figure*}

For transparency, we also provide qualitative examples of image synthesis failures that were filtered out during human validation (Figure~\ref{fig:app_qual_error}). These cases illustrate scenarios where the generated images are incomplete or misaligned with the given textual prompt. Based on our results, our synthetic images provide useful insights into the generation pipeline and can help guide future efforts in constructing culture-mixing datasets.

\begin{table*}[ht]
\footnotesize
  \centering
  \begin{tabular}{r l c p{0.6\linewidth}}
    \toprule
    idx & Country & Count & Food1 Names \\
    \midrule
    1 & United States Of America (USA) & 19 & 1. Baked beans; 2. Carolina-style pulled pork/barbecue; 3. Chocolate chip cookie; 4. co-shon duh lay; 5. Dutch letter; 6. Eggs Benedict; 7. Hamburger; 8. Jum-buh-lie-ahh; 9. Key Lime Pie; 10. Loco moco; 11. Mochi ice cream; 12. Pecan pie; 13. Pepper jelly; 14. Pie à la Mode; 15. Pumpkin pie; 16. Rainbow cookie; 17. Southern fried chicken; 18. Spaghetti and meatballs; 19. Tootsie Roll \\
    2 & China & 10 & 1. Century egg; 2. Cong you bing; 3. Edamame; 4. Lo mai gai; 5. Mantau; 6. Osmanthus cake; 7. Paper wrapped cake; 8. Yangzhou Fried Rice; 9. Yin yang fried rice; 10. Yong tau foo \\
    3 & France & 10 & 1. Beef bourguignon; 2. Bouneschlupp; 3. Chouquette; 4. Croissant; 5. Gâteau Basque; 6. Ladyfinger; 7. Nun's puffs; 8. Ratatouille; 9. Sablé; 10. Teurgoule \\
    4 & Indonesia & 10 & 1. Bakpia; 2. Kue mangkok; 3. Kue putu; 4. Laksa; 5. Lakso; 6. Nasi Uduk; 7. Sayur asem; 8. Siomay; 9. Tahu campur; 10. Tahu sumedang \\
    5 & Kenya & 10 & 1. Bacella alba; 2. Biegnets; 3. Cha-pa-ti or Cha-poh; 4. Githeri; 5. Karaage; 6. Kenyan kachumbari salad; 7. Maize and beans stew; 8. Matoke; 9. Mukimo; 10. Pound maize flour \\
    6 & Mexico & 10 & 1. Chilaquiles; 2. Chile relleno; 3. Huarache; 4. Migas; 5. Paste (pasty); 6. Piñata cookie; 7. Puchero; 8. Refried beans; 9. Salsa verde; 10. Sope \\
    7 & Philippines & 10 & 1. Aparon; 2. Beef Steak; 3. Laing; 4. Oyster omelette; 5. Puto; 6. Rendang; 7. Silog; 8. Sushi Bake; 9. Taho; 10. Uraro \\
    8 & Russia & 10 & 1. Alexandertorte; 2. Bracken fern salad; 3. Mimosa salad; 4. Potato pancake; 5. Pozharsky cutlet; 6. Shchi; 7. Solyanka; 8. Ukha; 9. Vinegret; 10. Zefir \\
    9 & South Africa & 10 & 1. African spinach; 2. Bunny chow; 3. Chicken and mushroom pie; 4. Hertzoggie; 5. kota, skhambane; 6. Malva pudding; 7. Melktert; 8. Mopane stew; 9. Potjiekos; 10. tripe \\
    10 & Spain & 10 & 1. Andrajos; 2. Arròs negre; 3. Cocido lebaniego; 4. Cocido madrileño; 5. Ensaïmada; 6. Escudella; 7. Hamin; 8. Hornazo; 9. Panellets; 10. Tortillitas de camarones \\
    11 & Italy & 9 & 1. Cannoli; 2. Casoncelli; 3. Cavallucci; 4. Cotoletta alla milanese; 5. Florentine biscuit; 6. Gelato; 7. Michetta; 8. Piadina romagnola; 9. Piccata \\
    12 & Japan & 9 & 1. Amanattō; 2. Char siu; 3. Dorayaki; 4. Fried ice cream; 5. Melonpan; 6. no; 7. Omurice; 8. Simmered dried strips of daikon radish; 9. Tempura \\
    13 & United Kingdom & 9 & 1. Bedfordshire clanger; 2. Blackberry pie; 3. Empire biscuit; 4. Ham and cheese sandwich; 5. Pease pudding; 6. Saveloy; 7. Spotted dick; 8. Stornoway black pudding; 9. Treacle sponge pudding \\
    14 & Egypt & 8 & 1. Custard tart; 2. Falafel; 3. Koshary; 4. Molokhia; 5. mombar; 6. Qatayef; 7. Um ali \\
    15 & India & 8 & 1. Bhel puri; 2. Curd rice; 3. Neer dosa; 4. Panta bhat; 5. Pulihora; 6. Rice and curry; 7. Thalassery biryani; 8. Unnakai \\
    16 & Korea & 8 & 1. Bibim guksu; 2. Bibimbap; 3. Japchae; 4. Jeon; 5. Milmyeon; 6. Sundubu-jjigae; 7. Tteokbokki; 8. Tteokguk \\
    17 & Nigeria & 8 & 1. Bridie; 2. Cooked cassava flakes and Okra soup; 3. E/ku/ru; 4. Ekwang; 5. Lupis; 6. Moin moin; 7. Okra Soup; 8. Vegetable soup with Egusi \\
    18 & Portugal & 8 & 1. Aletria; 2. Bacalhau à Gomes de Sá; 3. Cabidela; 4. Cebolada; 5. Mocotó; 6. Pastel de Tentúgal; 7. Polvo à lagareiro; 8. Tripas à moda do Porto \\
    19 & Algeria & 7 & 1. Algerian Almond Cookies; 2. Algerian Mekrout; 3. Besbousa; 4. Chermoula; 5. Harira; 6. Tamina; 7. Tikerbabin \\
    20 & Germany & 7 & 1. Edi-kang Ikong; 2. Eisbein; 3. Kaiserschmarrn; 4. Maultasche; 5. Poppy seed roll; 6. Toast Hawaii \\
    21 & Thailand & 7 & 1. Khanom bueang; 2. Khao kan chin; 3. Khao soi; 4. Kluai khaek; 5. Mi krop; 6. Pad thai; 7. Sakhu sai mu \\
    22 & England & 6 & 1. Bacon and egg pie; 2. Bakewell tart; 3. Bath bun; 4. Curry pie; 5. Fish pie; 6. Steak pie \\
    23 & Iran & 6 & 1. Baghali polo; 2. Chelow; 3. Kashk bademjan; 4. Mirza ghassemi; 5. Reshteh khoshkar; 6. Sajji \\
    24 & Cameroon & 5 & 1. Achu; 2. leaves of gnetum; 3. Plantain Chips; 4. Puff Puff, beans and pape; 5. rice with beans \\
    25 & Croatia & 5 & 1. Brudet; 2. Grahova pretepena juha; 3. Međimurska gibanica; 4. soparnik; 5. Zagorski štrukli \\
    26 & Nepal & 5 & 1. Chataamari; 2. Chhurpi; 3. Gajar ka halwa; 4. Kwati; 5. Sapu Mhicha \\
    27 & Pakistan & 5 & 1. Amba; 2. Bun kebab; 3. Momo; 4. Phirni; 5. Zarda \\
    28 & Singapore & 5 & 1. Banmian; 2. Bihun Goreng; 3. Chwee kueh; 4. Noodles with tomato egg sauce; 5. Turnip cake \\
    29 & Turkey & 5 & 1. Bamia; 2. Cezerye; 3. Kuru fasulye; 4. Qurabiya; 5. Şöbiyet \\
    30 & Poland & 4 & 1. cabbage rolls; 2. Dumplings; 3. Gołąbki; 4. Krumiri \\
    31 & Sudan & 4 & 1. Cucumber salad with yogurt; 2. Khachapuri; 3. LoQeymat or zalabia; 4. Vanille cake \\
    \bottomrule
  \end{tabular}
  \caption{\textbf{Unique food counts and names per country included in \task.}}
  \label{tab:food_stats}
\end{table*}

\onecolumn
\begin{tcolorbox}[
    colback=gray!5,
    colframe=gray!30,
    coltitle=black, 
    boxrule=0.3pt,
    width=\textwidth,
    arc=0mm, 
    title=Prompts for \SF Image Generation,
    fonttitle=\bfseries,
]
\begin{itemize}
    \item Leave all the food quality the exact same as the original.
    \item Modify the background image to a pure white.
    \item Make the image square. Change the food into top-down view.
    \item Convert the food to a top-down view.
    \item Remove any spoons, chopsticks, and human hands. 
    \item Add any missing parts of a plate or a bowl. 
    \item The plate or bowl should be circle or oval.
\end{itemize}
\end{tcolorbox}

\begin{tcolorbox}[
    colback=gray!5,
    colframe=gray!30,
    coltitle=black, 
    boxrule=0.3pt,
    width=\textwidth,
    arc=0mm, 
    title=Prompts for \SFB Image Generation (FLUX),
    fonttitle=\bfseries
]
\begin{itemize}
    \item Change the white background underneath a single food item to a table or picnic mat.
    \item The table or picnic mat should seamlessly be integrated with the background image.
    \item Rotate the food items along the z-axis so they are viewed from a natural dining perspective --- not from the top, but more like how someone sees the plate while sitting at a table, and add realistic shadows to the plates.
    \item Remove any hands and utensils on the plates.
    \item Reconstruct the plate if there isn't any or if it's broken.
    \item Keep the background and the food quality identical as the original.
    \item The generated image should have only one food item.
\end{itemize}
\end{tcolorbox}

\begin{tcolorbox}[
    colback=gray!5,
    colframe=gray!30,
    coltitle=black, 
    boxrule=0.3pt,
    width=\textwidth,
    arc=0mm, 
    title=Prompts for \SFB Image Generation (QWEN),
    fonttitle=\bfseries
]
\begin{itemize}
    \item Leave all the elements the exact same as the original except for the followings:
    \item Add a table or picnic mat underneath the food item.
    \item The table or picnic mat should seamlessly be integrated with the background image.
    \item Rotate the food items along the z-axis so they are viewed from a natural dining perspective --- not from the top, but more like how someone sees the plate while sitting at a table, and add realistic shadows to the plates.
    \item Remove any hands and utensils on the plates.
    \item Reconstruct the plate if there isn't any or if it's broken.
    \item The generated image should have only one food item.
\end{itemize}
\end{tcolorbox}

\begin{tcolorbox}[
    colback=gray!5,
    colframe=gray!30,
    coltitle=black, 
    boxrule=0.3pt,
    width=\textwidth,
    arc=0mm, 
    title=Prompts for \MFB Image Generation (FLUX),
    fonttitle=\bfseries
]
\begin{itemize}
    \item Change the white background underneath two food items to a table or picnic mat.
    \item The table or picnic mat should seamlessly be integrated with the background image.
    \item Rotate the food items along the z-axis so they are viewed from a natural dining perspective --- not from the top, but more like how someone sees the plate while sitting at a table, and add realistic shadows to the plates.
    \item Remove any hands and utensils on the plates.
    \item Reconstruct the plate if there isn't any or if it's broken.
    \item Keep the background and the food quality identical as the original.
    \item The generated image should have only two food items.
\end{itemize}
\end{tcolorbox}

\begin{tcolorbox}[
    colback=gray!5,
    colframe=gray!30,
    coltitle=black, 
    boxrule=0.3pt,
    arc=0mm, 
    title=Prompts for \MFB Image Generation (QWEN),
    fonttitle=\bfseries
]
\begin{itemize}
    \item Leave all the elements the exact same as the original except for the following:
    \item Add a table or picnic mat underneath the food item.
    \item The table or picnic mat should seamlessly be integrated with the background image.
    \item Rotate the food items along the z-axis so they are viewed from a natural dining perspective --- not from the top, but more like how someone sees the plate while sitting at a table, and add realistic shadows to the plates.
    \item Remove any hands and utensils on the plates.
    \item Reconstruct the plate if there isn't any or if it's broken.
    \item The generated image should have only two food items.
\end{itemize}
\end{tcolorbox}
\twocolumn
\begin{table}[h!]
\centering
\caption{\textbf{Models and Configurations.} Overview of the models and their specific settings used for image synthesis.}
\resizebox{\linewidth}{!}{
\begin{tabular}{ll}
\toprule
\textbf{Category} & \textbf{Details} \\ \hline
\textbf{Models} & \begin{tabular}[l]{@{}l@{}}
\texttt{black-forest-labs/FLUX.1-Kontext-dev}\\
\texttt{Qwen/Qwen-Image-Edit}
\end{tabular} \\ \midrule
\textbf{Configs} & \begin{tabular}[l]{@{}l@{}}
\texttt{guidance\_scale = 2.5}\\
\texttt{num\_inference\_steps = 50}\\
\texttt{torch.bfloat16}
\end{tabular} \\
\bottomrule
\end{tabular}
}
\label{tab:model-config}
\end{table}

\subsection{Sample Dataset Images}
\label{appendix:sample_imgs}
\begin{figure*}[ht]

        \centering
        \includegraphics[width=\textwidth]{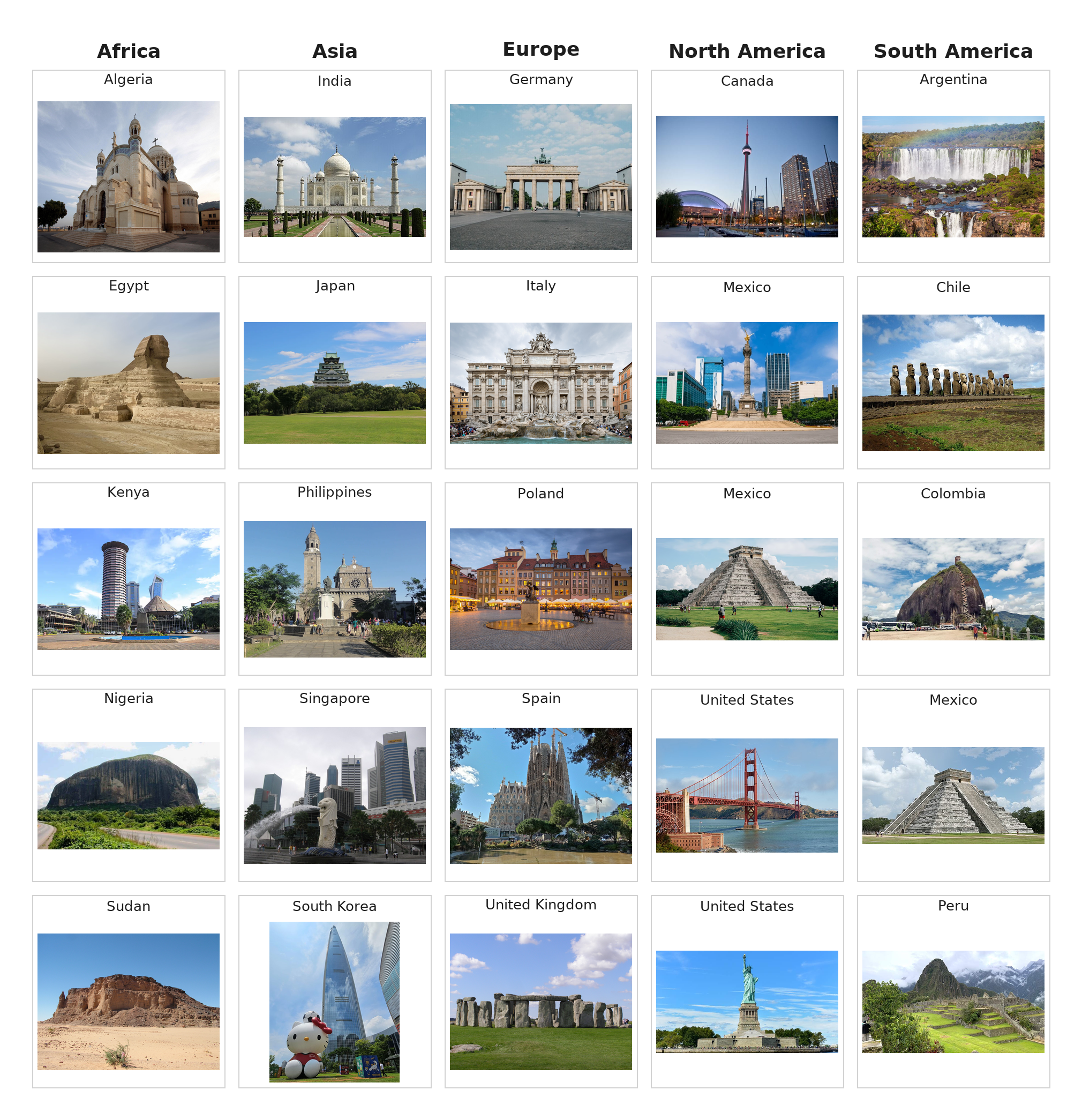}
       \caption{\textbf{Background (Landmark) Images.}}
        \label{fig:landmark_bgs}
\end{figure*}

\begin{figure*}[ht]
        \centering
        \includegraphics[width=\textwidth]{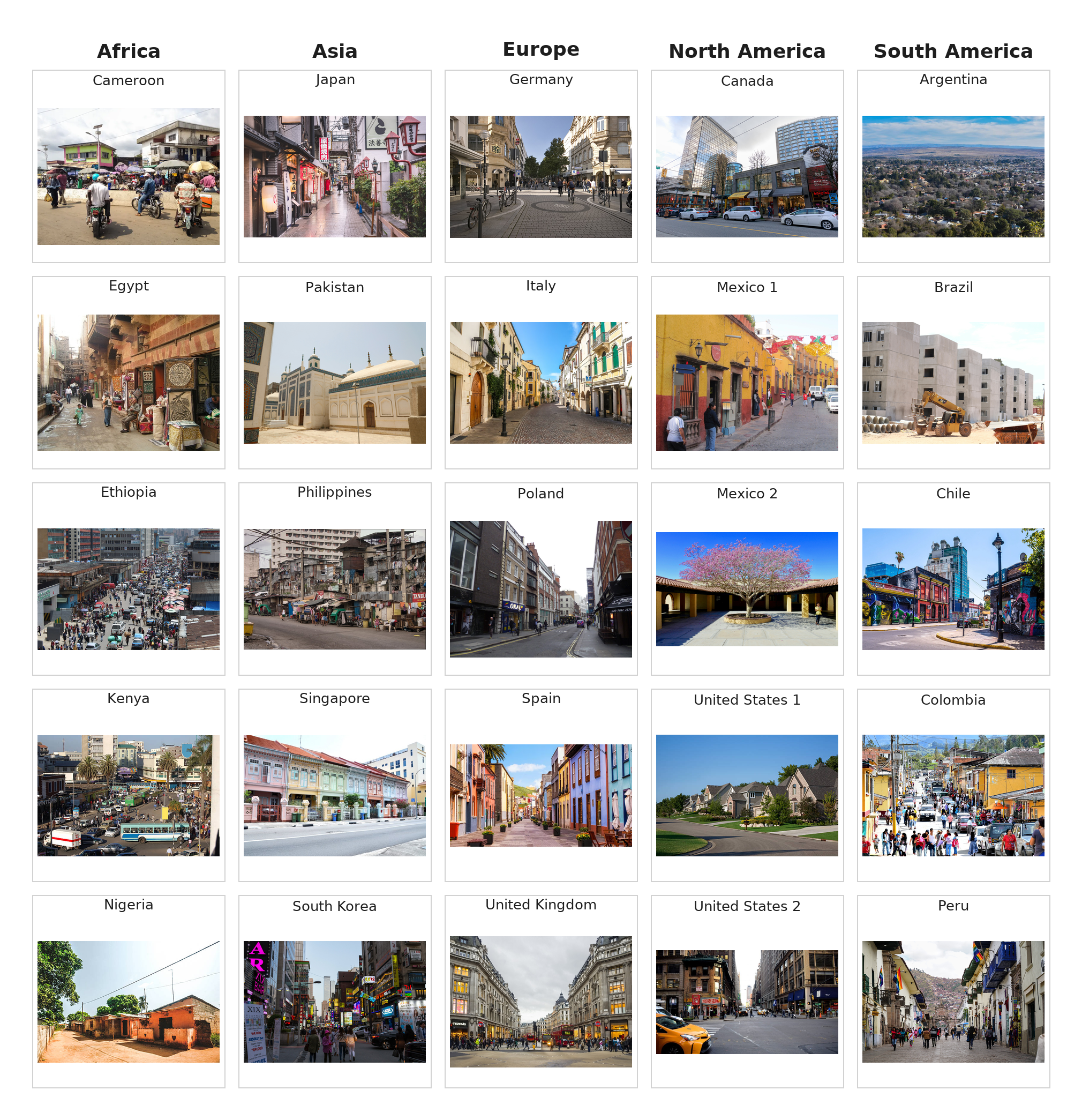}
        \caption{\textbf{Background (Street) Images.}}
        \label{fig:street_bgs}
\end{figure*}

The complete background image set, consisting of 25 landmark images and 25 street images, is shown in Figure~\ref{fig:landmark_bgs} and Figure~\ref{fig:street_bgs}, respectively.
\paragraph{Final image generation samples}

\begin{figure*}[ht]
    \centering
    \includegraphics[width=\textwidth]{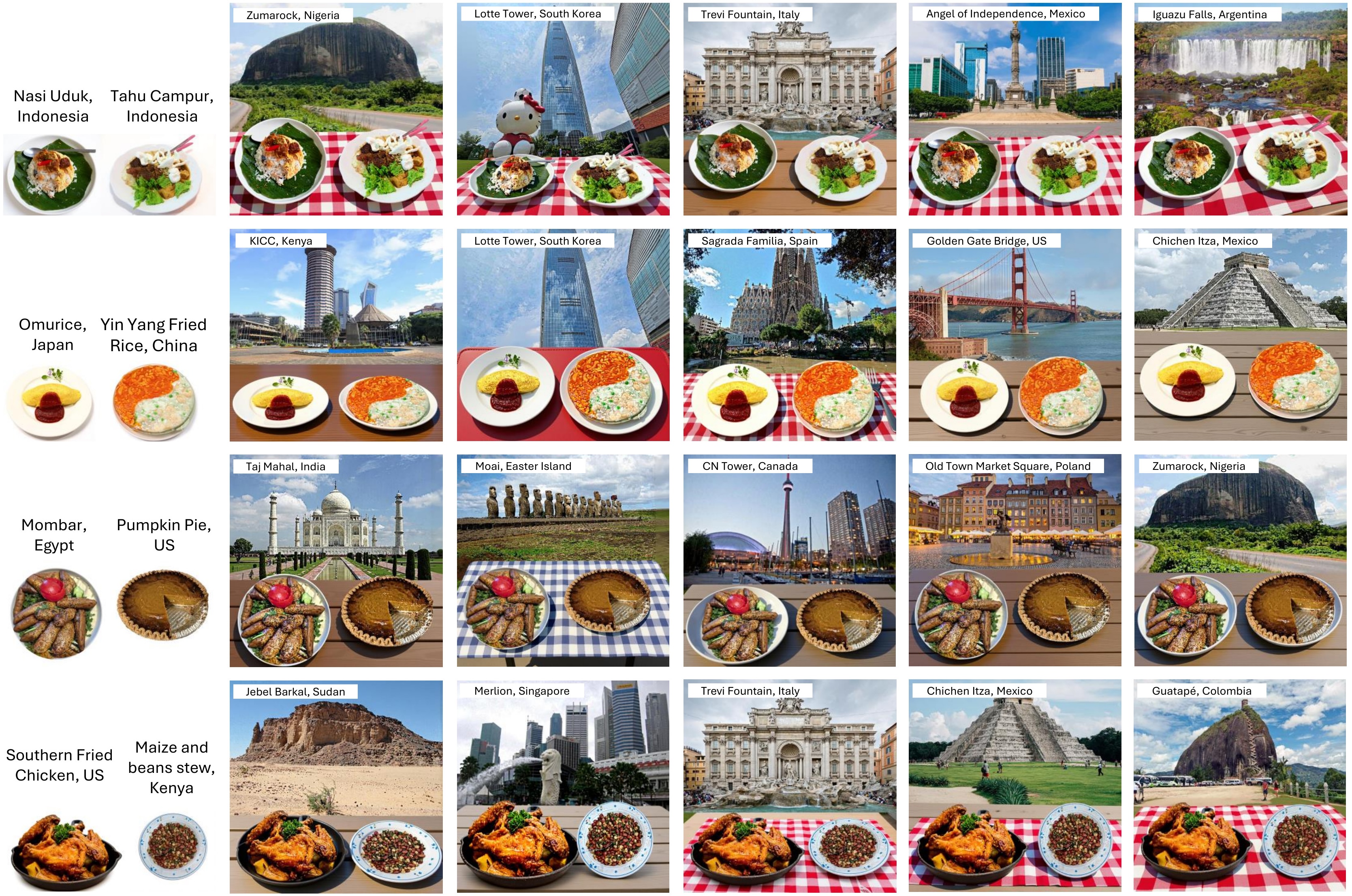}
    \caption{\textbf{Image Generation Examples.} Each row of composite images shows each food from different countries placed in multiple diverse global backgrounds. Results illustrate the cultural combinations represented in our dataset.} 
    \label{fig:synthetic}
\end{figure*}

Figure~\ref{fig:synthetic} presents examples of generated culture-mixing images from our constructed dataset. These samples illustrate the diversity of cross-cultural compositions produced during the construction of our dataset, illustrating variations in food types and cultural geographic backgrounds. These images reflect the range of visual configurations used to evaluate how LVLMs interpret cultural signals when multiple cultural elements coexist within a single image. 

\paragraph{Real-world image samples}

\begin{figure*}[ht]
    \centering
    \includegraphics[width=0.8\textwidth]{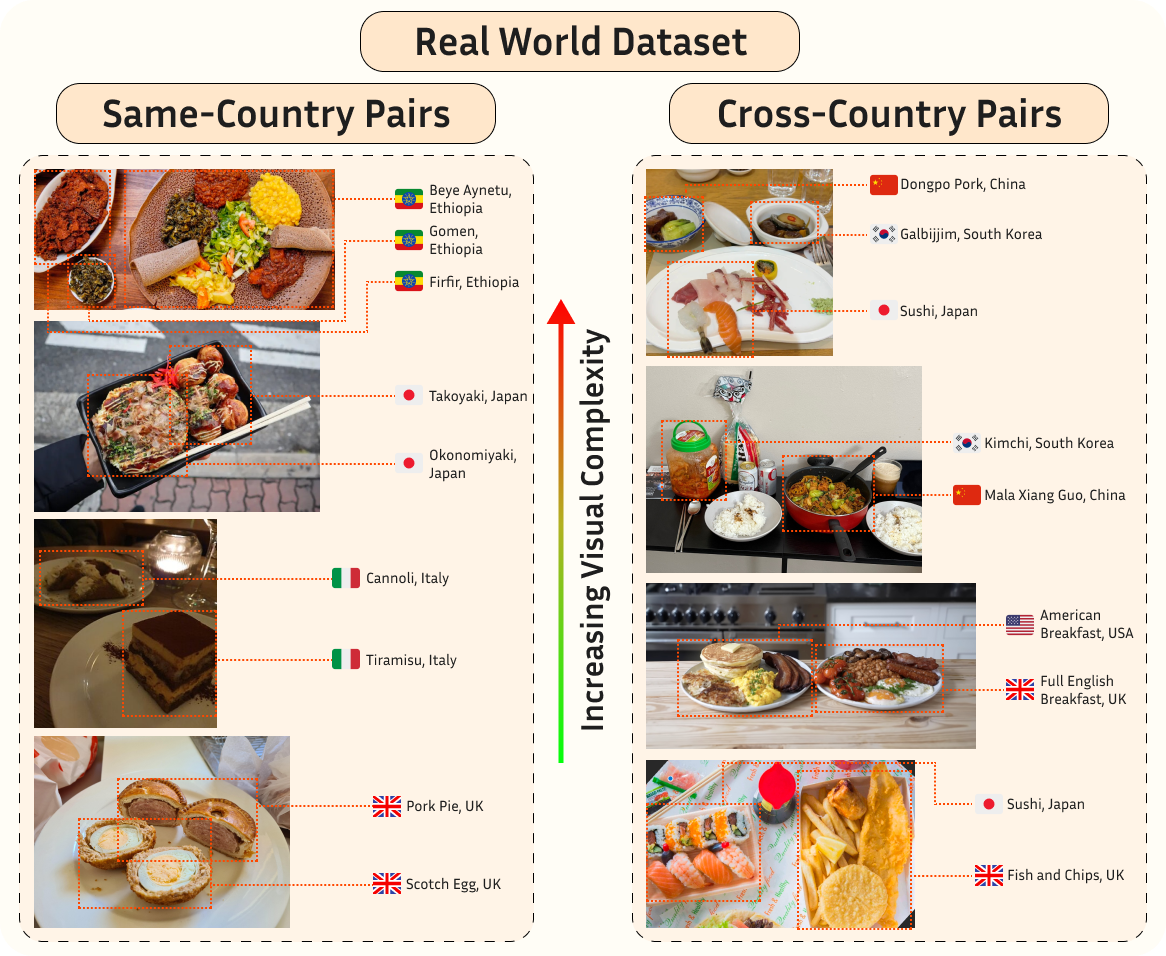}
    \caption{\textbf{Real World Dataset Example.} Same-country food pairs (left) and cross-country pairs (right) are shown across a spectrum of visual complexity, reflecting the diversity of appearance, plating, and scene structure encountered in real images } 
    \label{fig:realworld}
\end{figure*}

To complement generated images, we also collect a real-world image set featuring naturally occurring food pairings (Figure~\ref{fig:realworld}). We categorize half of the collected images into same-country pairs and the other into cross-country pairs. Images were also collected based on a variety of visual complexity, reflecting common sources of variation in real photographs such as mixed lighting, cluttered environments, irregular plating, and inconsistent camera perspectives.




\clearpage
\section{Experiments}
\subsection{Results}
\label{appendix:results}
\paragraph{Country and Food Name Identification Accuracy}

Table~\ref{tab:supp_country_name_acc} reports each model's country and food name identification accuracy across subtasks, providing additional detail that complements the radar chart in Figure~3a in the main text.
Figure~\ref{fig:app_sankey} visualizes how prediction correctness shifts from \SF to the mixed subtasks (\SFB, \MF, \MFB), showing that culturally mixed contexts often cause models to fail on cases they initially predicted correctly.
Also, Figure~\ref{fig:app_country_heatmap} repeats Figure~3b in the main text with a larger size for closer inspection.
Additionally, Table~\ref{tab:supp_background_type_compare} compares the models' country and food name identification accuracy across landmark and street backgrounds. The performance difference between these two background types was minimal for both \SFB and \MFB.
Figure~\ref{fig:cultural_distance_analysis} compares the country, food name identification accuracy, and entropy according to different cultural distances between the target and the distractor, which complements Figure~5 in the main text.

\paragraph{Country and Food Name Identification Entropy}
Table~\ref{tab:country_food_entropy_unnorm} reports each model's country and food name identification entropy across subtasks, providing additional detail that complements Figure~6 in the main text.

\paragraph{Real World Dataset}

Table~\ref{tab:real_country_name_acc} reports each model's country and food name identification accuracy of the real-world dataset, providing additional detail that complements the radar chart in Figure~8 in the main text.


\subsection{Ablations}
\label{appendix:bias_ablations}
\paragraph{Positional Bias}
To examine the effect of positional bias with respect to food location on the model performance, we randomly sample 100 multi-food images and compare the predicted labels before and after horizontally flipping each image. As shown in Table~\ref{tab:locbias_name}, the high consistency between the original and flipped predictions indicates that the model's outputs are stable under left–right reversals, suggesting minimal to no positional dependence for food-related attributes.


\paragraph{Size Bias}
We also investigate the effect of the food item sizes on the model performance by first comparing the relative sizes of food items appearing on the left and right within multi-food images and then evaluating whether differing size ratios lead to changes in the predicted labels. Table~\ref{tab:posbias_name} demonstrates that the model exhibits minimal size-related bias when identifying food items. In other words, even when one food item is noticeably larger than the other, the model's identification accuracy remains stable, suggesting that its predictions are largely invariant to object size differences.


\newpage

\subsection{Qualitative Analysis on Food Name Prediction Failure Cases}
\begin{figure}[!h]
    \centering
    \includegraphics[width=\linewidth]{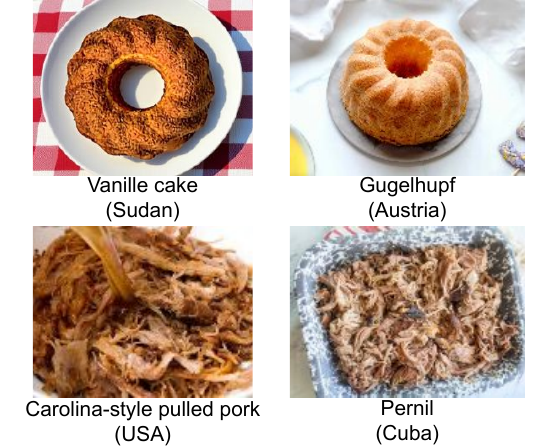}
    \caption{\textbf{Similar Food Prediction Examples.} Gemini predicted \textit{Vanille cake} as \textit{Gugelhupf}, and InternVL3-8B predicted \textit{Carolina-style pulled pork} as \textit{Pernil}. In both cases, the predicted dishes are visually similar to the ground-truth foods yet are distinct items.}
    \label{fig:similar_food_pred}
\end{figure}

We sampled 25 instances of incorrect food-name predictions from each subtask (\SF, \MF, \SFB, \MFB) for Gemini-2.5-pro and InternVL3-8B, resulting in 100 images per model. We then manually checked whether the model's predicted food label visually resembled the ground-truth food (i.e., the model confused it with a similar-looking dish) or whether it was entirely unrelated, using Google Image search results for the predicted food as reference.

Gemini-2.5-pro predicted a visually similar food in 25 out of 100 cases, whereas InternVL3-8B did so in only 2 out of 100. This indicates that both models often predict completely different foods, but between the two, Gemini-2.5-pro tended to make closer guesses, whereas InternVL3-8B's predictions showed little resemblance to the target food. Figure~\ref{fig:similar_food_pred} shows examples of similar food prediction for each model.


\clearpage
\newpage

\begin{table*}[ht]
\centering
\caption{\textbf{Identification accuracy performance.} Most LVLMs show relatively higher accuracy in single food settings (SFB and SF) compared to that of multiple food settings (MFB and MF) for both country and food name identification. \textbf{Bold} and \underline{underline} indicate the highest and lowest accuracy among subtasks, respectively.}

\begin{subtable}[t]{\linewidth}
    \centering
    \caption{Country Identification}
    \begin{tabular}{lcccc}
        \toprule
        \textbf{Model} & \textbf{SF (N=988)} & \textbf{MF (N=948)} & \textbf{SFB (N=12,350)} & \textbf{MFB (N=9,485)} \\
        \midrule
        Gemini-2.5-Pro & 0.457 & \textbf{0.499} & \underline{0.286} & 0.313 \\
        GPT-5 & \textbf{0.487} & 0.450 & \underline{0.250} & 0.271 \\
        InternVL3-78B & \textbf{0.234} & 0.231 & \underline{0.110} & 0.125 \\
        InternVL3-38B & \textbf{0.205} & 0.199 & \underline{0.088} & 0.112 \\
        InternVL3-14B & 0.161 & \textbf{0.170} & \underline{0.075} & 0.087 \\
        InternVL3-8B & \textbf{0.152} & 0.140 & \underline{0.065} & 0.071 \\
        Molmo-72B & \textbf{0.139} & 0.129 & \underline{0.080} & 0.081 \\
        QwenVL3-32B & \textbf{0.242} & 0.213 & \underline{0.077} & 0.124 \\
        QwenVL3-8B & \textbf{0.252} & 0.227 & \underline{0.060} & 0.124 \\
        Ovis-9B & \textbf{0.285} & 0.263 & \underline{0.133} & 0.148 \\
        \midrule
        \textbf{Avg.} & \textbf{0.261} & 0.252 & \underline{0.122} & 0.146 \\
        \bottomrule
    \end{tabular}
    \label{tab:country_acc}
\end{subtable}

\vspace{1.5em}

\begin{subtable}[t]{\textwidth}
    \centering
    \caption{Name Identification}
    \begin{tabular}{lcccc}
        \toprule
        \textbf{Model} & \textbf{SF (N=988)} & \textbf{MF (N=948)} & \textbf{SFB (N=12,350)} & \textbf{MFB (N=9,485)} \\
        \midrule
        Gemini-2.5-Pro & 0.399 & \textbf{0.435} & \underline{0.252} & 0.268 \\
        GPT-5 & \textbf{0.379} & 0.371 & \underline{0.199} & 0.212 \\
        InternVL3-78B & \textbf{0.128} & 0.113 & \underline{0.065} & 0.069 \\
        InternVL3-38B & \textbf{0.100} & 0.093 & \underline{0.056} & 0.062 \\
        InternVL3-14B & \textbf{0.076} & 0.071 & \underline{0.043} & 0.045 \\
        InternVL3-8B & 0.070 & \textbf{0.072} & \underline{0.035} & 0.040 \\
        Molmo-72B & \textbf{0.090} & \textbf{0.093} & 0.067 & \underline{0.058} \\
        QwenVL3-32B & \textbf{0.164} & 0.079 & \underline{0.045} & 0.084 \\
        QwenVL3-8B & \textbf{0.145} & 0.129 & \underline{0.034} & 0.071 \\
        Ovis-9B & \textbf{0.152} & 0.112 & 0.077 & \underline{0.075} \\
        \midrule
        \textbf{Avg.} & \textbf{0.170} & 0.157 & \underline{0.087} & 0.098 \\
        \bottomrule
    \end{tabular}
    \label{tab:name_acc}
\end{subtable}
\label{tab:supp_country_name_acc}

\end{table*}

\begin{figure*}[ht]
    \centering
    \includegraphics[width=\textwidth]{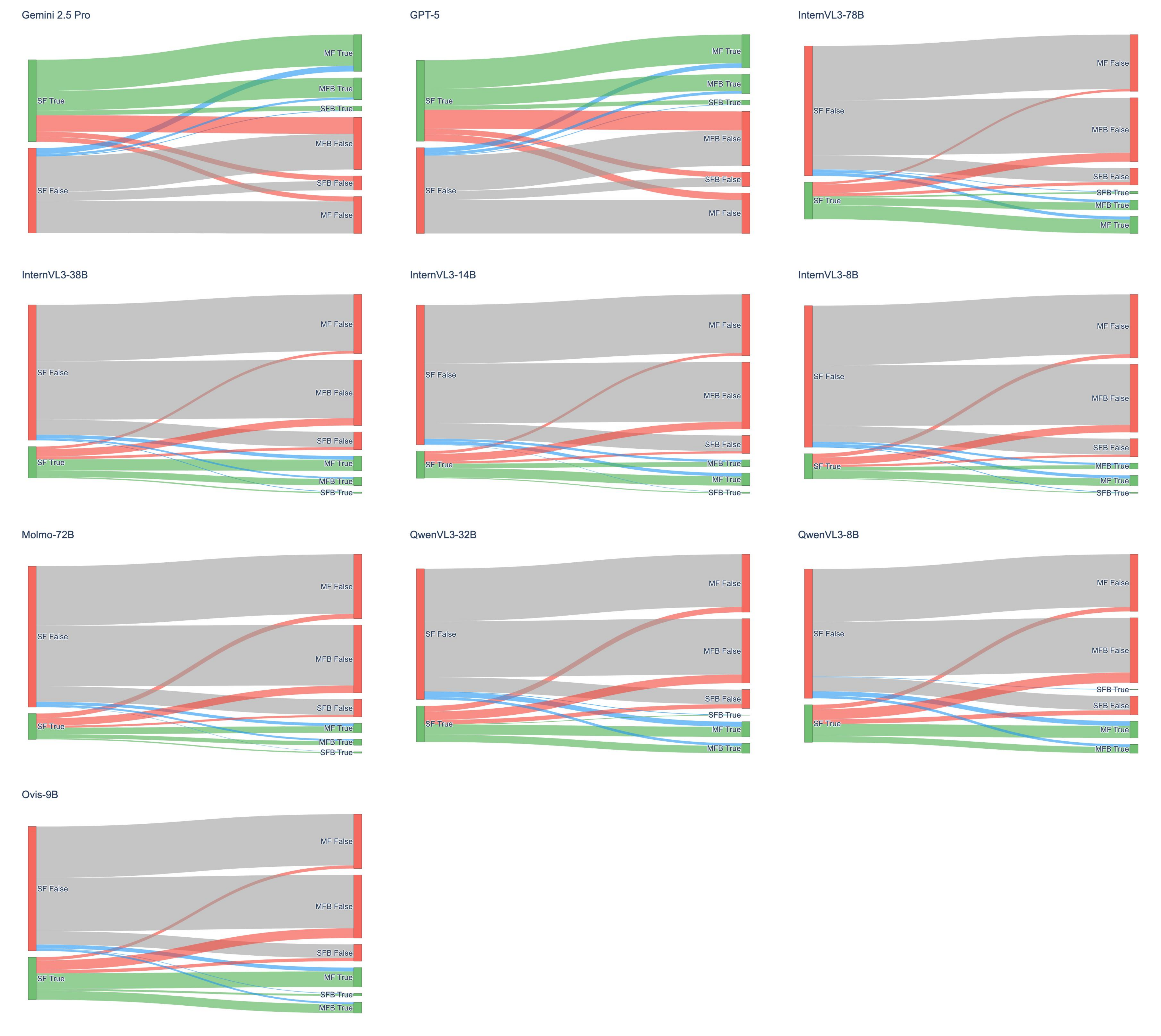}
    \caption{\textbf{Sankey diagram of country prediction for each LVLM.} We visualize how prediction correctness shifts from \SF to the mixed subtasks (\SFB, \MF, \MFB). Green indicates True $\rightarrow$ True, Red indicates True $\rightarrow$ False, Blue indicates False $\rightarrow$ True, and Gray indicates False $\rightarrow$ False. Although a small portion of predictions fall into Blue (False $\rightarrow$ True), a much larger portion appears in Red (True $\rightarrow$ False), showing that culturally mixed contexts confuse the models and often cause them to fail on cases they initially predicted correctly. While closed-source models perform better on country identification in \SF, they also exhibit substantial True $\rightarrow$ False shifts in the mixed subtasks, resulting in reduced accuracy in culturally mixed settings.}
    \label{fig:app_sankey}
\end{figure*}

\begin{figure*}[ht]
    \centering
    \includegraphics[width=\textwidth]{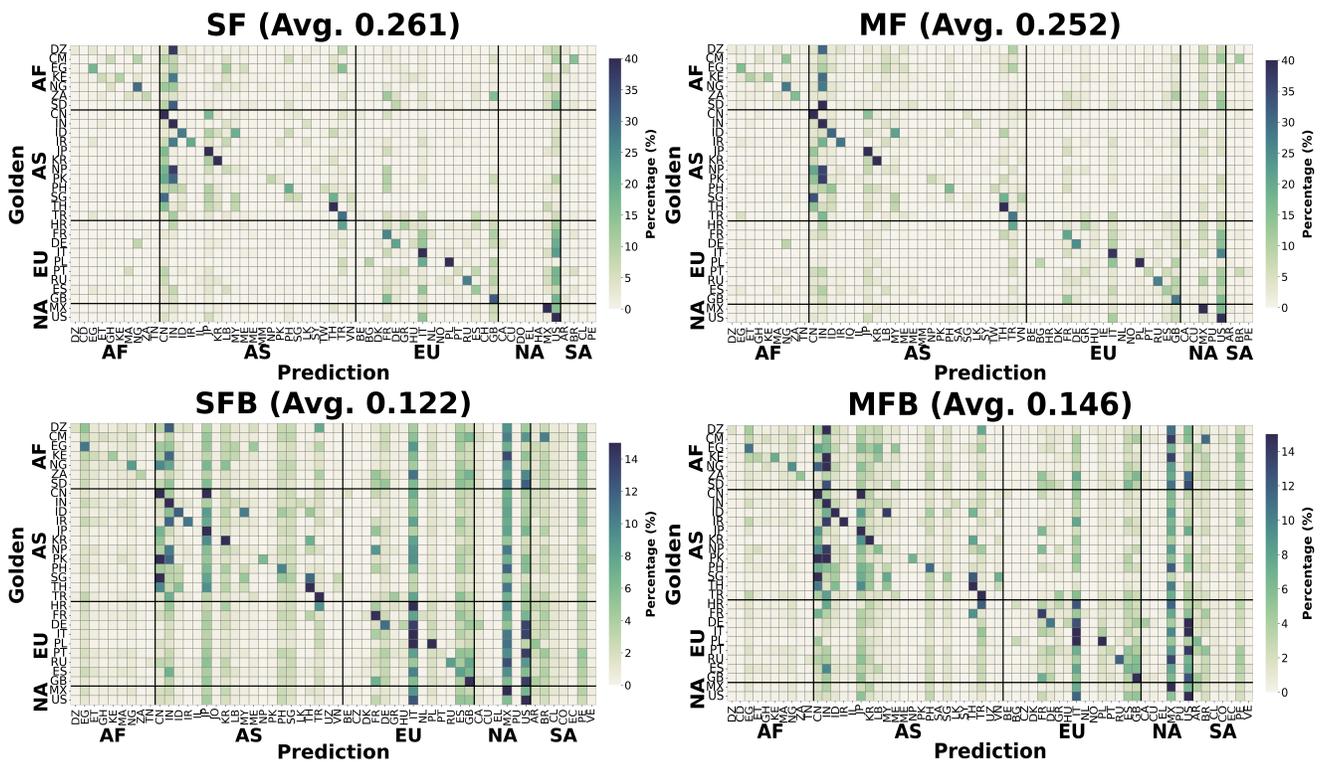}
    \caption{\textbf{Country identification target-prediction heatmaps for each subtask.} For every golden country, the plots show the distribution
of predicted countries, illustrating both correct predictions and systematic confusions across models. The figure is repeated here in a larger size to facilitate closer examination.}
\label{fig:app_country_heatmap}
\end{figure*}

\begin{table*}[ht]
\centering
\caption{\textbf{Country and food name identification accuracy by background (\SFB and \MFB).} The models perform similarly for identifying the country and food with the landmark and street background.}

\begin{subtable}[t]{\linewidth}
    \centering
    \caption{\SFB Accuracy by Background Type}
    \begin{tabular}{lcccc}
        \toprule
        \multirow{2}{*}{\textbf{Model}} & 
        \multicolumn{2}{c}{\textbf{Country}} & 
        \multicolumn{2}{c}{\textbf{Food Name}} \\
        \cmidrule(lr){2-3} \cmidrule(lr){4-5}
         & \textbf{Landmark} & \textbf{Street} & \textbf{Landmark} & \textbf{Street} \\
        \midrule
        Gemini-2.5-Pro   & 0.285 & 0.288 & 0.255 & 0.250 \\
        GPT-5            & 0.249 & 0.252 & 0.200 & 0.199 \\
        InternVL3-78B    & 0.119 & 0.101 & 0.068 & 0.063 \\
        InternVL3-38B    & 0.086 & 0.089 & 0.057 & 0.055 \\
        InternVL3-14B    & 0.078 & 0.073 & 0.044 & 0.041 \\
        InternVL3-8B     & 0.067 & 0.063 & 0.037 & 0.033 \\
        Molmo-72B        & 0.080 & 0.081 & 0.067 & 0.068 \\
        QwenVL3-32B      & 0.033 & 0.122 & 0.003 & 0.086 \\
        QwenVL3-8B       & 0.028 & 0.092 & 0.000 & 0.067 \\
        Ovis-9B          & 0.139 & 0.126 & 0.080 & 0.075 \\
        \midrule
        \textbf{Avg.}    & 0.116 & 0.129 & 0.081 & 0.094 \\
        \bottomrule
    \end{tabular}
    \label{tab:sfb_background}
\end{subtable}

\par\vspace{1.5em}

\begin{subtable}[t]{\linewidth}
    \centering
    \caption{\MFB Accuracy by Background Type}
    \begin{tabular}{lcccc}
        \toprule
        \multirow{2}{*}{\textbf{Model}} & 
        \multicolumn{2}{c}{\textbf{Country}} & 
        \multicolumn{2}{c}{\textbf{Food Name}} \\
        \cmidrule(lr){2-3} \cmidrule(lr){4-5}
         & \textbf{Landmark} & \textbf{Street} & \textbf{Landmark} & \textbf{Street} \\
        \midrule
        Gemini-2.5-Pro   & 0.311 & 0.314 & 0.264 & 0.272 \\
        GPT-5            & 0.265 & 0.277 & 0.204 & 0.220 \\
        InternVL3-78B    & 0.134 & 0.117 & 0.071 & 0.068 \\
        InternVL3-38B    & 0.116 & 0.107 & 0.064 & 0.060 \\
        InternVL3-14B    & 0.095 & 0.080 & 0.045 & 0.044 \\
        InternVL3-8B     & 0.076 & 0.067 & 0.039 & 0.040 \\
        Molmo-72B        & 0.082 & 0.080 & 0.057 & 0.059 \\
        QwenVL3-32B      & 0.133 & 0.115 & 0.088 & 0.079 \\
        QwenVL3-8B       & 0.129 & 0.118 & 0.072 & 0.070 \\
        Ovis-9B          & 0.156 & 0.139 & 0.078 & 0.072 \\
        \midrule
        \textbf{Avg.}    & 0.150 & 0.141 & 0.098 & 0.098 \\
        \bottomrule
    \end{tabular}
    \label{tab:mfb_background}
\end{subtable}
\label{tab:supp_background_type_compare}

\end{table*}


\begin{figure*}[ht]
    \centering

    \begin{subfigure}{0.9\textwidth}
        \centering
        \includegraphics[width=\linewidth]{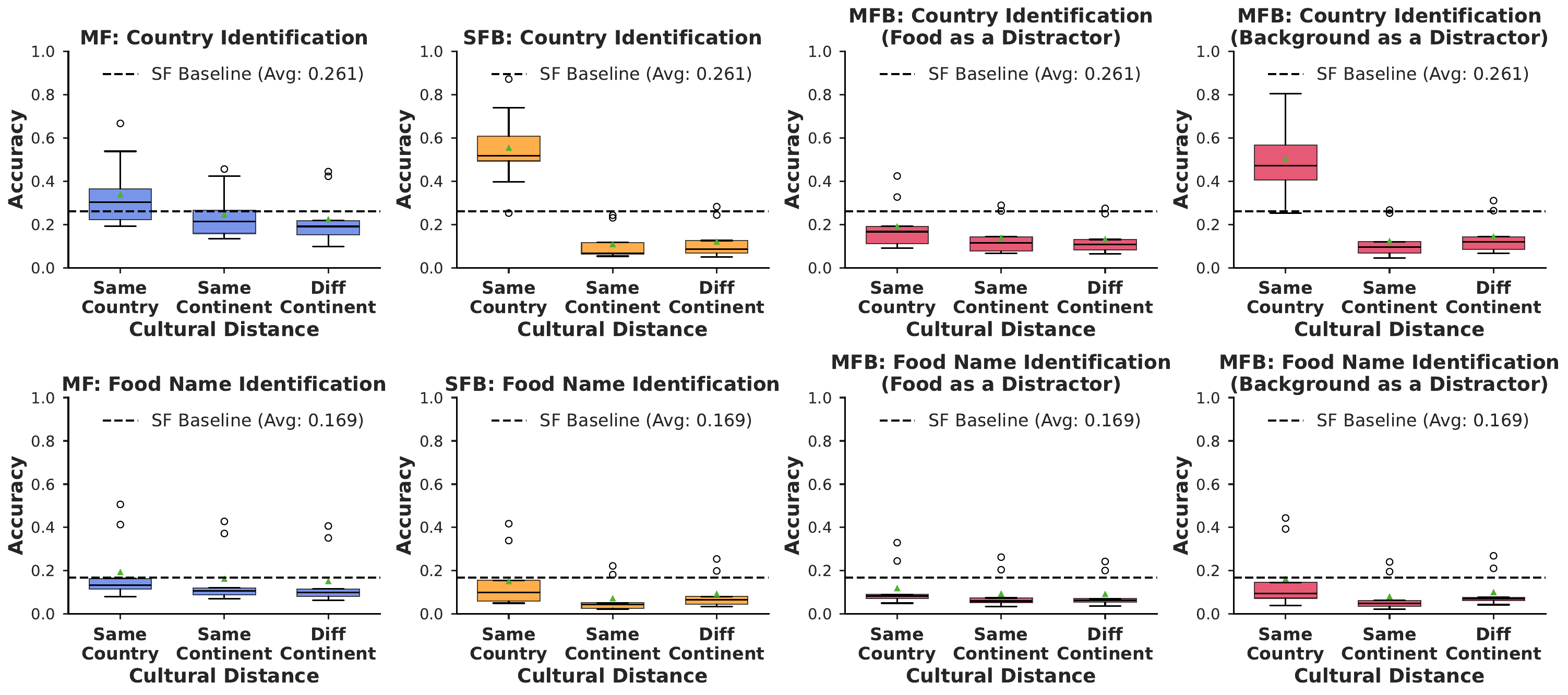}
        \caption{Cultural distance Vs. Accuracy}
        \label{fig:cultural_distance_accuracy}
    \end{subfigure}

    \vspace{0.5em} 

    \begin{subfigure}{0.9\textwidth}
        \centering
        \includegraphics[width=\linewidth]{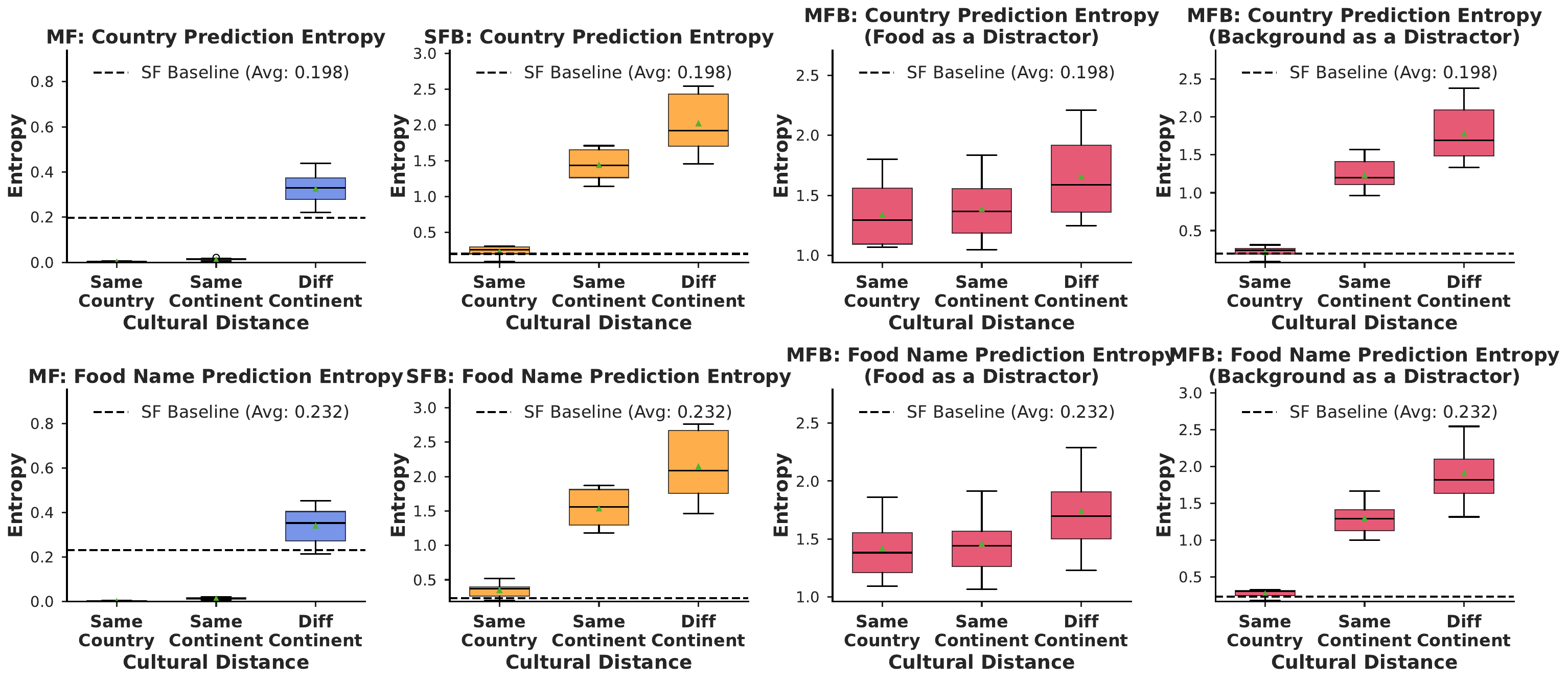}
        \caption{Cultural distance Vs. Entropy}
        \label{fig:cultural_distance_entropy}
    \end{subfigure}

    \caption{\textbf{Effect of cultural distance between target and distractor.}
    Models perform best when the target and distractor originate from the same country. Accuracy declines and entropy increases as cultural distance increases, indicating room for improving culture-mixing understanding in LVLMs.}
    \label{fig:cultural_distance_analysis}
\end{figure*}

\begin{table*}[ht]
\centering
\caption{\textbf{Predicted label entropy by model and subtask.} The  LVLMs show relatively high prediction uncertainty in single food settings (SFB and SF) compared to that of multiple food settings (MFB and MF) for both country and food name identification. \textbf{Bold} and \underline{underline} indicate the highest and lowest accuracy among subtasks, respectively.}

\begin{subtable}[t]{\linewidth}
    \centering
    \caption{Country Identification Entropy}
    \begin{tabular}{lccccc}
        \toprule
        \textbf{Model} & \textbf{SF} & \textbf{MF} & \textbf{SFB} & \textbf{MFB} & \textbf{Object} \\
        \midrule
        Gemini-2.5-Pro & \underline{0.1674} & 0.4466 & \textbf{1.9906} & 1.7028 & 0.3127 \\
        GPT-5 & \underline{0.1406} & 0.3774 & \textbf{1.6294} & 1.4498 & 0.4073 \\
        InternVL3-78B & \underline{0.1545} & 0.5471 & \textbf{2.0483} & 1.7942 & 0.4895 \\
        InternVL3-38B & \underline{0.1358} & 0.5294 & \textbf{1.5570} & 1.4269 & 0.5136 \\
        InternVL3-14B & \underline{0.2109} & 0.6882 & \textbf{2.5861} & 2.3096 & 0.7084 \\
        InternVL3-8B & \underline{0.2279} & 0.7483 & \textbf{2.6767} & 2.5273 & 0.7784 \\
        Molmo-72B & \underline{0.3794} & 0.8183 & 1.7660 & \textbf{1.9765} & 1.0114 \\
        QwenVL3-32B & \underline{0.2124} & \textbf{1.0243} & 2.7046 & 1.7793 & 0.6309 \\
        QwenVL3-8B & \underline{0.1894} & 0.6087 & \textbf{2.5548} & 1.5627 & 0.6015 \\
        Ovis-9B & \underline{0.1628} & 0.7224 & \textbf{2.4157} & 2.2161 & 0.4315 \\
        \midrule
        \textbf{Average} & \underline{0.1981} & 0.6511 & \textbf{2.1929} & 1.8745 & 0.5831 \\
        \bottomrule
    \end{tabular}
    \label{tab:country_entropy_unnorm}
\end{subtable}

\vspace{1.5em}

\begin{subtable}[t]{\linewidth}
    \centering
    \caption{Food Name Identification Entropy}
    \begin{tabular}{lccccc}
        \toprule
        \textbf{Model} & \textbf{SF} & \textbf{MF} & \textbf{SFB} & \textbf{MFB} & \textbf{Object} \\
        \midrule
        Gemini-2.5-Pro & \underline{0.7720} & 0.9071 & \textbf{2.2122} & 2.0513 & 0.9076 \\
        GPT-5 & 0.8014 & \underline{0.8144} & \textbf{1.9083} & 1.7651 & 0.8574 \\
        InternVL3-78B & \underline{0.7264} & 0.9045 & \textbf{2.2624} & 1.9843 & 0.8797 \\
        InternVL3-38B & \underline{0.7515} & 0.8252 & \textbf{1.5961} & 1.4558 & 0.8136 \\
        InternVL3-14B & \underline{0.7840} & 0.9458 & \textbf{2.8754} & 2.5452 & 0.9444 \\
        InternVL3-8B & \underline{0.8079} & 0.9485 & \textbf{2.9280} & 2.7138 & 0.9707 \\
        Molmo-72B & \underline{0.9270} & 1.0470 & 1.7244 & \textbf{2.1658} & 1.1527 \\
        QwenVL3-32B & \underline{0.7933} & \textbf{1.1436} & 3.1793 & 2.0625 & 0.9798 \\
        QwenVL3-8B & \underline{0.7914} & 0.9590 & \textbf{2.7944} & 1.6854 & 0.9032 \\
        Ovis-9B & \underline{0.7685} & 1.0390 & \textbf{2.3858} & 2.2773 & 0.9155 \\
        \midrule
        \textbf{Average} & \underline{0.7923} & 0.9534 & \textbf{2.3866} & 2.0707 & 0.9325 \\
        \bottomrule
    \end{tabular}
    \label{tab:food_entropy_unnorm}
\end{subtable}
\label{tab:country_food_entropy_unnorm}
\end{table*}

\begin{table*}[ht]
\centering
\caption{\textbf{Real world dataset identification accuracy.} LVLMs generally show better performance in identifying multiple foods from the same culture, even in real-world settings.}
\begin{subtable}[t]{\linewidth}
    \centering
    \caption{Country Identification Accuracy}
    \begin{tabular}{lccc}
        \toprule
        \textbf{Model} & \textbf{Single} & \textbf{Multi (Same)} & \textbf{Multi (Diff)} \\
        \midrule
        Gemini-2.5-Pro & 0.868 & \textbf{0.922} & \underline{0.856} \\
        GPT-5 & 0.877 & \textbf{0.904} & \underline{0.750} \\
        InternVL3-78B & 0.721 & \textbf{0.809} & \underline{0.654} \\
        InternVL3-38B & \underline{0.667} & \textbf{0.791} & 0.683 \\
        InternVL3-14B & 0.562 & \textbf{0.609} & \underline{0.558} \\
        InternVL3-8B & \underline{0.566} & \textbf{0.678} & 0.663 \\
        Molmo-72B & 0.539 & \textbf{0.670} & \underline{0.500} \\
        QwenVL3-32B & \textbf{0.658} & 0.600 & \underline{0.356} \\
        QwenVL3-8B & 0.731 & \textbf{0.748} & \underline{0.375} \\
        Ovis-9B & 0.749 & \textbf{0.896} & \underline{0.692} \\
        \midrule
        \textbf{Average} & 0.694 & \textbf{0.763} & \underline{0.609} \\
        \bottomrule
    \end{tabular}
    \label{tab:country_multi_all}
    
\vspace{1.5em}   

\begin{subtable}[t]{\linewidth}
    \centering
    \caption{Name Identification Accuracy}
    \begin{tabular}{lccc}
        \toprule
        \textbf{Model} & \textbf{Single} & \textbf{Multi (Same)} & \textbf{Multi (Diff)} \\
        \midrule
        Gemini-2.5-Pro & 0.699 & \textbf{0.765} & \underline{0.692} \\
        GPT-5 & \textbf{0.635} & 0.617 & \underline{0.519} \\
        InternVL3-78B & 0.511 & \textbf{0.583} & \underline{0.471} \\
        InternVL3-38B & \underline{0.443} & \textbf{0.513} & 0.471 \\
        InternVL3-14B & \underline{0.365} & \textbf{0.400} & 0.375 \\
        InternVL3-8B & \underline{0.384} & 0.391 & \textbf{0.413} \\
        Molmo-72B & 0.311 & \textbf{0.365} & \underline{0.298} \\
        QwenVL3-32B & \textbf{0.489} & 0.478 & \underline{0.327} \\
        QwenVL3-8B & \textbf{0.530} & 0.461 & \underline{0.337} \\
        Ovis-9B & 0.511 & \textbf{0.548} & \underline{0.500} \\
        \midrule
        \textbf{Average} & 0.488 & \textbf{0.512} & \underline{0.440} \\
        \bottomrule
    \end{tabular}
    \label{tab:name_multi_all}
\end{subtable}
\end{subtable}
\label{tab:real_country_name_acc}
\end{table*}

\begin{table*}[ht]
\centering
\caption{\textbf{The effect of food item location on identification accuracy.} We observe almost no shift when comparing predictions on the original multi-food images and their horizontally flipped counterparts, indicating minimal positional bias in both country and name identification.}
\begin{subtable}[t]{\linewidth}
    \centering
    \caption{Country Identification Accuracy}
    \begin{tabular}{lccc}
        \toprule
        \textbf{Model} & \textbf{MF (N=350)} & \textbf{Location Shift (N=100)} & \textbf{Diff (Location $-$ MF)} \\
        \midrule
        InternVL3-8B & 0.111 & 0.090 & $-$0.021 \\
        Ovis-9B & 0.237 & 0.250 & $+$0.013 \\
        QwenVL3-8B & 0.203 & 0.180 & $-$0.023 \\
        \bottomrule
    \end{tabular}
    \label{tab:locbias_country}
\end{subtable}

\vspace{1.5em}

\begin{subtable}[t]{\linewidth}
    \centering
    \caption{Name Identification Accuracy}
    \begin{tabular}{lccc}
        \toprule
        \textbf{Model} & \textbf{MF (N=350)} & \textbf{Location Shift (N=100)} & \textbf{Diff (Location $-$ MF)} \\
        \midrule
        InternVL3-8B & 0.049 & 0.020 & $-$0.029 \\
        Ovis-9B & 0.086 & 0.100 & $+$0.014 \\
        QwenVL3-8B & 0.131 & 0.120 & $-$0.011 \\
        \bottomrule
    \end{tabular}
\end{subtable}
\label{tab:locbias_name}
\end{table*}

\begin{table*}[ht]
\centering
\caption{\textbf{The effect of size on identification accuracy.} We observe almost no change in model performance between the original single-food images and their resized variants, indicating that identification accuracy is largely invariant to object size.}
\begin{subtable}[t]{\linewidth}
    \centering
    \caption{Country Identification}
    \begin{tabular}{lccc}
        \toprule
        \textbf{Model} & \textbf{SF (N=247)} & \textbf{Size Shift (N=741)} & \textbf{Diff} \\
        \midrule
        InternVL3-8B & 0.158 & 0.139 & $-$0.019 \\
        Ovis-9B & 0.271 & 0.290 & $+$0.019 \\
        QwenVL3-8B & 0.239 & 0.224 & $-$0.015 \\
        \bottomrule
    \end{tabular}
    \label{tab:posbias_country}
\end{subtable}
\par\vspace{1.5em}
\begin{subtable}[t]{\linewidth}
    \centering
    \caption{Food Name Identification}
    \begin{tabular}{lccc}
        \toprule
        \textbf{Model} & \textbf{SF (N=247)} & \textbf{Size Shift (N=741)} & \textbf{Diff} \\
        \midrule
        InternVL3-8B & 0.061 & 0.066 & $+$0.005 \\
        Ovis-9B & 0.150 & 0.152 & $+$0.003 \\
        QwenVL3-8B & 0.142 & 0.143 & $+$0.001 \\
        \bottomrule
    \end{tabular}
\end{subtable}
\label{tab:posbias_name}
\end{table*}


\end{document}